\documentclass[]{uiuc_paper}
\pdfoutput=1

\usepackage[toc,page,header]{appendix}
\usepackage{cleveref}
\usepackage{subcaption}
\usepackage{booktabs}
\usepackage{tabularx}
\usepackage{array}
\usepackage{colortbl}
\usepackage{algorithm}
\usepackage{algpseudocode}
\usepackage{tikz}
\usetikzlibrary{arrows.meta,positioning,fit,calc,shapes.geometric,backgrounds}

\usepackage{latexsym}
\usepackage{url}
\usepackage{amssymb}
\usepackage{amsmath}
\usepackage{amsfonts}
\usepackage[utf8]{inputenc}
\usepackage{booktabs}
\usepackage{pifont}
\usepackage{multirow}
\usepackage{makecell}
\usepackage{paralist}
\usepackage{xspace}
\usepackage{adjustbox}

\crefname{section}{Sec.}{Secs.}\Crefname{section}{Sec.}{Secs.}
\crefname{subsection}{Sec.}{Secs.}\Crefname{subsection}{Sec.}{Secs.}
\crefname{subsubsection}{Sec.}{Secs.}\Crefname{subsubsection}{Sec.}{Secs.}
\crefname{appendix}{App.}{Apps.}\Crefname{appendix}{App.}{Apps.}
\crefname{subappendix}{App.}{Apps.}\Crefname{subappendix}{App.}{Apps.}
\crefname{subsubappendix}{App.}{Apps.}\Crefname{subsubappendix}{App.}{Apps.}
\crefname{figure}{Fig.}{Figs.}\Crefname{figure}{Fig.}{Figs.}
\crefname{table}{Tab.}{Tabs.}\Crefname{table}{Tab.}{Tabs.}
\crefname{equation}{Eq.}{Eqs.}\Crefname{equation}{Eq.}{Eqs.}
\crefname{algorithm}{Alg.}{Algs.}\Crefname{algorithm}{Alg.}{Algs.}

\newcommand{\method}{Vanguard}

\definecolor{vgnavy}{HTML}{2F4B8F}
\definecolor{vgblue}{HTML}{8797B0}
\definecolor{vgmist}{HTML}{E6EBF4}
\definecolor{vgrose}{HTML}{C29C9B}
\definecolor{vgtext}{HTML}{2B3F5B}

\title{Vanguard: Knowing When to Trust the Draft in Speculative Vision-Language Agents}

\author[1,*]{Hanqi Mao}
\author[1,*]{Jiajun Fan}
\author[1]{Ruihan Guo}
\author[2]{Shuaike Shen}
\author[1,3]{Ziwen Wang}
\author[1]{Hui Ren}
\author[4]{Xianrui Zhong}
\author[1]{Ge Liu}

\affiliation[1]{\affillogo{\includegraphics{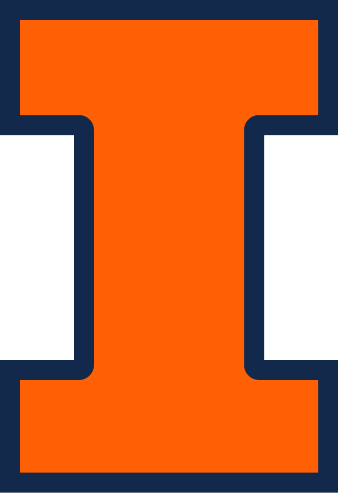}}{UIUC}}
\affiliation[2]{\affillogo{\includegraphics{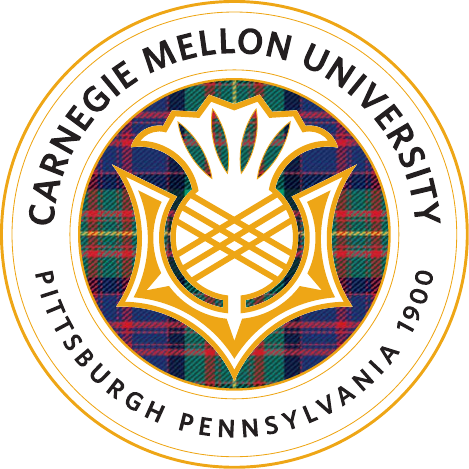}}{CMU}}
\affiliation[3]{\affillogo{\includegraphics{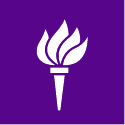}}{NYU}}
\affiliation[4]{\affillogo{\includegraphics{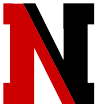}}{Northeastern}}

\contribution[*]{Equal contribution}

\abstract{Vision-language agents that browse websites, operate phones and act in homes reason before every action, so every step waits for a large model, a latency and cost that stand in the way of deployment.
Step-level speculation lets a small model draft each step and calls the large model only on rejected drafts, but its outcome rests on the judge, and a judge that reads only a step's text cannot tell whether the step fits the screen it acts on.
We find that a few samples of a small model can contain a correct step about as often as the large model gets the step right, and not always on the same steps: checked against the screen, small drafts can be not only cheaper but better.
We present \method{}, a training-free judge that leaves the draft and target models unchanged: a general decision model, with the screenshot in view, gives each candidate step a probability of being right in one forward pass; the best candidate is kept if it clears a threshold, and otherwise the large model writes the step.
On web, mobile and embodied agents, a 2B draft judged by \method{} is at least as accurate as its 8B target and significantly more accurate on Mind2Web and EB-ALFRED, while the target writes only a minority of the steps; it completes far more closed-loop episodes, and with a 27B target from another family it is both faster and more accurate.
\method{} is the most accurate judge we compare on the agent tasks; it relies on the screenshot on web and mobile tasks, its calibrated threshold transfers to unseen websites, and across model families its gain over the target grows with the draft's pass@4 relative to the target's accuracy.
\method{} composes with token-level speculative decoding and matches trained process reward models on mathematics, turning speculation from a way to save the large model's work into one that can improve on it.
}

\date{\today}

\begin{document}

\maketitle

\section{Introduction}
\label{sec:intro}
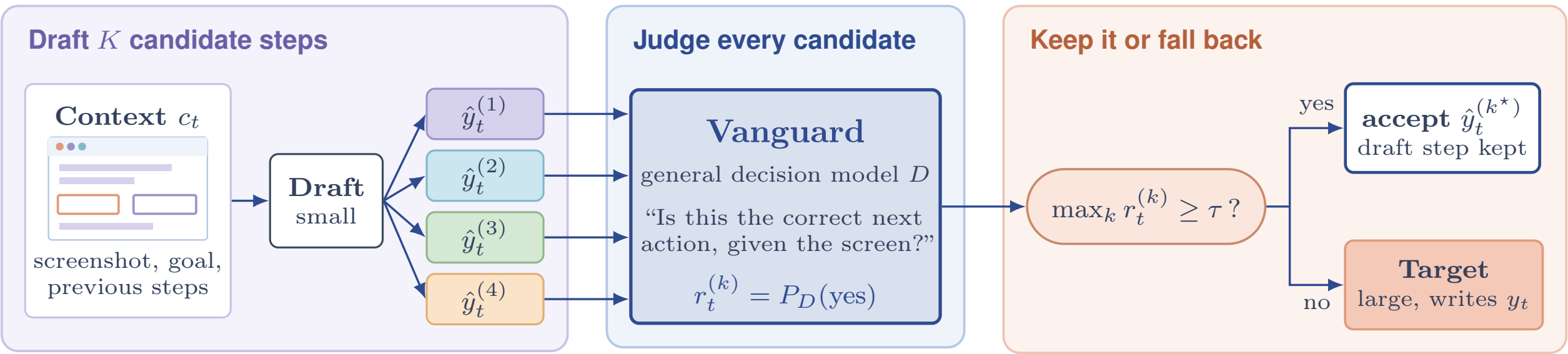
\begin{figure}[t]
\centering
\definecolor{fNavy}{HTML}{2F4B8F}\definecolor{fText}{HTML}{33415C}
\definecolor{fOursF}{HTML}{D9E0EE}\definecolor{fDraftE}{HTML}{3D4A5C}
\definecolor{fTgtF}{HTML}{F5C9B7}\definecolor{fTgtE}{HTML}{E19A7E}
\definecolor{fLavF}{HTML}{D6D1EC}\definecolor{fLavE}{HTML}{9C93C9}\definecolor{fCyanF}{HTML}{CBE6EE}\definecolor{fCyanE}{HTML}{7FBACB}
\definecolor{fGreenF}{HTML}{D5E8D4}\definecolor{fGreenE}{HTML}{8DB98A}
\definecolor{fAprF}{HTML}{FBE3C8}\definecolor{fAprE}{HTML}{E2AE72}
\definecolor{fDecF}{HTML}{FAE6DC}\definecolor{fDecE}{HTML}{C98A6D}
\definecolor{pLav}{HTML}{F4F2FA}\definecolor{pBlue}{HTML}{EFF4FA}\definecolor{pPeach}{HTML}{FCF3EE}
\definecolor{bLav}{HTML}{C9C3E4}\definecolor{bBlue}{HTML}{B3C8E6}\definecolor{bPeach}{HTML}{EDC3B2}
\definecolor{tLav}{HTML}{6A5FA6}\definecolor{tBlue}{HTML}{2F4B8F}\definecolor{tPeach}{HTML}{B4603D}
\renewcommand{\normalsize}{\fontsize{10}{12}\selectfont}\renewcommand{\footnotesize}{\fontsize{8}{9.5}\selectfont}\renewcommand{\scriptsize}{\fontsize{7}{8}\selectfont}%
\resizebox{\linewidth}{!}{%
\begin{tikzpicture}[
  font=\footnotesize, text=fText, >={Latex[length=1.7mm,width=1.4mm]},
  box/.style={rounded corners=2.5pt, align=center, line width=0.6pt},
  chip/.style={box, minimum width=12.5mm, minimum height=5.4mm, inner sep=1pt},
  arr/.style={->, draw=fNavy, line width=0.65pt},
  lab/.style={font=\scriptsize, text=fText},
  panel/.style={rounded corners=5pt, line width=0.7pt, inner sep=2.4mm},
  ptitle/.style={font=\footnotesize\bfseries\sffamily, anchor=north west}]
  \node[box, draw=bLav, fill=white, minimum width=22mm, minimum height=25mm] (ctx) {};
  \node[font=\footnotesize\bfseries, anchor=north] at ([yshift=-1.2mm]ctx.north) {Context $c_t$};
  \begin{scope}[shift={([yshift=-5.8mm]ctx.north)}]
    \draw[bBlue, fill=white, rounded corners=1pt, line width=0.45pt] (-8.5mm,-11mm) rectangle (8.5mm,0);
    \fill[pBlue] (-8.5mm,-2mm) rectangle (8.5mm,0);
    \foreach \x/\c in {-7.3/fTgtE,-6.1/fLavE,-4.9/fCyanE} \fill[\c] (\x mm,-1mm) circle (0.42mm);
    \foreach \y/\w in {-3.6/12,-5.0/8,-9.9/10} \fill[fLavF] (-7.3mm,\y mm) rectangle ++(\w mm,0.7mm);
    \draw[fTgtE, line width=0.7pt, rounded corners=0.6pt] (-7.5mm,-8.2mm) rectangle (-1mm,-6.2mm);
    \draw[fLavE, line width=0.7pt, rounded corners=0.6pt] (0.6mm,-8.2mm) rectangle (7.3mm,-6.2mm);
  \end{scope}
  \node[lab, anchor=south, align=center] at ([yshift=0.8mm]ctx.south) {screenshot, goal,\\previous steps};
  \node[box, draw=fDraftE, fill=white, text width=11mm, minimum height=10mm, inner sep=1.5pt, right=4mm of ctx] (draft) {\textbf{Draft}\\[-1pt]{\scriptsize small}};
  \node[chip, draw=fLavE, fill=fLavF, right=4.5mm of draft, yshift=9.3mm] (c1) {$\hat y^{(1)}_t$};
  \node[chip, draw=fCyanE, fill=fCyanF, below=1.0mm of c1] (c2) {$\hat y^{(2)}_t$};
  \node[chip, draw=fGreenE, fill=fGreenF, below=1.0mm of c2] (c3) {$\hat y^{(3)}_t$};
  \node[chip, draw=fAprE, fill=fAprF, below=1.0mm of c3] (c4) {$\hat y^{(4)}_t$};
  \node[fit=(c1)(c4), inner sep=0pt] (cands) {};
  \node[box, draw=fNavy, fill=fOursF, line width=1.0pt, text=fText, right=9mm of cands, minimum height=25mm, text width=31mm, inner sep=3pt] (judge) {};
  \node[text=fNavy, font=\normalsize\bfseries, anchor=north] at ([yshift=-2mm]judge.north) {\method{}};
  \node[text=fText, font=\scriptsize, anchor=north] at ([yshift=-7mm]judge.north) {general decision model $D$};
  \node[text=fText, align=center, anchor=north, font=\scriptsize, text width=31mm] at ([yshift=-11.6mm]judge.north) {``Is this the correct next\\action, given the screen?''};
  \node[text=fNavy, anchor=north] at ([yshift=-18.6mm]judge.north) {$r^{(k)}_t = P_D(\text{yes})$};
  \node[box, rounded corners=4mm, minimum height=8mm, inner xsep=2.6mm, right=9mm of judge, draw=fDecE, fill=fDecF, line width=0.7pt] (dec) {$\max_k r^{(k)}_t \ge \tau$\,?};
  \node[box, draw=fNavy, fill=white, line width=0.9pt, minimum width=20mm, minimum height=9.5mm, anchor=west] (keep) at ([xshift=8.3mm, yshift=8.4mm]dec.east) {\textbf{accept} $\hat y^{(k^\star)}_t$\\[-1pt]{\scriptsize draft step kept}};
  \node[box, draw=fTgtE, fill=fTgtF, line width=0.7pt, minimum width=20mm, minimum height=9.5mm, anchor=west] (tgt) at ([xshift=8.3mm, yshift=-8.4mm]dec.east) {\textbf{Target}\\[-1pt]{\scriptsize large, writes $y_t$}};
  \draw[arr] (ctx.east |- draft) -- (draft.west);
  \foreach \c in {c1,c2,c3,c4} \draw[arr] (draft.east) -- (\c.west);
  \foreach \c in {c1,c2,c3,c4} \draw[arr] (\c.east) -- (judge.west |- \c.east);
  \draw[arr] (judge.east |- dec) -- (dec.west);
  \coordinate (fork) at ([xshift=2.5mm]dec.east);
  \draw[draw=fNavy, line width=0.65pt] (dec.east) -- (fork);
  \draw[arr] (fork) |- node[pos=0.75, above=0.1mm, lab]{yes} (keep.west);
  \draw[arr] (fork) |- node[pos=0.75, below=0.1mm, lab]{no} (tgt.west);
  \coordinate (top) at ([yshift=6.4mm]judge.north);
  \coordinate (bot) at ([yshift=-0.4mm]judge.south);
  \begin{scope}[on background layer]
    \node[panel, draw=bLav, fill=pLav, fit=(ctx)(cands)(ctx.west |- top)(ctx.west |- bot)] (pA) {};
    \node[panel, draw=bBlue, fill=pBlue, fit=(judge)(judge.north |- top)] (pB) {};
    \node[panel, draw=bPeach, fill=pPeach, fit=(dec)(keep)(tgt)(keep.north |- top)(tgt.south |- bot)] (pC) {};
  \end{scope}
  \node[ptitle, text=tLav] at ([xshift=1.6mm,yshift=-1.3mm]pA.north west) {Draft $K$ candidate steps};
  \node[ptitle, text=tBlue] at ([xshift=1.6mm,yshift=-1.3mm]pB.north west) {Judge every candidate};
  \node[ptitle, text=tPeach] at ([xshift=1.6mm,yshift=-1.3mm]pC.north west) {Keep it or fall back};
\end{tikzpicture}}
\caption{\textbf{\method{} keeps a drafted step only when a decision model, looking at the screenshot, judges it right.} The small draft proposes $K = 4$ candidate steps; a general decision model scores each action against the screenshot in one forward pass; the best is kept if it clears $\tau$, otherwise the target writes the step.}
\label{fig:pipeline}
\end{figure}

Vision-language agents now browse websites, operate phones and act in households by reading a screenshot or a camera image, reasoning about the goal, and choosing an action, one step after another.
The agents that work best reason at length before every action~\cite{yang2026languagemodelsgaintesttime}, so each step is a full decoding pass of a large model: the better the agent, the more every step costs, and this latency and cost stand in the way of deploying it.
Speculative decoding~\cite{leviathan2023fast,chen2023accelerating} offers a way out: a small draft model proposes, and the large model is consulted only when needed.
At the token level, the draft is checked against the large model's own distribution, which keeps the output exact but requires the large model to score every token.
Step-level methods relax this requirement~\cite{liao2025rsd,pan2025specreason,liu2026confspec}: a small model drafts a whole step, a judge scores the draft, and the large model writes the step only when the judge rejects it.
Everything then depends on the judge: it decides what is saved, what is lost, and, as this paper shows, whether the small model can end up ahead of the large one.

The judges in use today fall into three groups, and each has a cost.
Some ask the large model to grade the draft~\cite{pan2025specreason,hu2026dreamr}, which turns generation into verification but still has the large model read every drafted step.
Some score the draft with a process reward model~\cite{liao2025rsd}, which is specialised to one domain and must be trained for it.
Others trust signals from the small model itself, such as its confidence or its attention~\cite{liu2026confspec,purohit2026tokens}, and inherit whatever the small model fails to notice about its own mistakes.

Visual inputs make the judgment harder.
A step can be fluent and logically sound, yet click an element that is not on the screen, or read the wrong value off a chart.
Token-level methods built for text-only models degrade on multimodal inputs~\cite{shen2026mmspec}, and a judge that sees only the text of a step cannot tell whether the step agrees with the image.

Our starting point is an observation about the draft: sampled a few times, a small vision-language model can propose a correct step about as often as the large model answers correctly (\cref{sec:exp_rule}), and not always on the same steps.
Existing judges use it poorly: the large model asked to check a draft puts almost all of its probability near zero or one (\cref{sec:s_where}), and the small model often misses its own mistakes.
What is missing is a calibrated judge that checks each candidate against the screen.

We present \method{}, such a judge for step-level speculation in vision-language agents.
At each step, the small model drafts several candidate steps, as in guided speculative inference~\cite{geuter2026gsi}.
For each candidate, \method{} asks a general decision model~\cite{typesafe2026jev} a yes-or-no question about whether the step is right, with the task and the screenshot in view, and reads off the probability of ``yes'' in a single forward pass, without generating any text.
The highest-scoring candidate is kept if its probability clears a threshold, and otherwise the large model writes the step (\cref{fig:pipeline}).
\method{} is training-free: only the wording of the question and the threshold are chosen on development data.
Unlike token-level speculation, it needs no draft heads trained for the target and no tokenizer shared with it, so the target may come from another model family; rather than reproducing the target's output, \method{} can improve on it, and the two levels of speculation compose.

We evaluate \method{} on web, mobile and embodied agents (Multimodal-Mind2Web~\cite{deng2023mind2web,zheng2024seeact}, AndroidControl~\cite{li2024androidcontrol} and EB-ALFRED~\cite{yang2025embodiedbench}), on closed-loop episodes of BabyAI~\cite{chevalierboisvert2019babyai}, and on visual and text mathematical reasoning (MathVista~\cite{lu2024mathvista}, MATH-500~\cite{hendrycks2021math,lightman2023verify} and ProcessBench~\cite{zheng2025processbench}).

Our contributions are:
\begin{itemize}
\item \textbf{Speculation that improves on the large model.} With four candidates and fallback, a 2B draft judged by \method{} is at least as accurate as its 8B target on every agent task, significantly more accurate on web and embodied tasks, and completes far more closed-loop episodes, while the target writes only a minority of the steps; with a 27B target from another family it is both faster and more accurate than the target, and it composes with token-level speculative decoding.
\item \textbf{A training-free judge that checks each drafted step against the screen.} \method{} leaves the draft and target models unchanged and asks a general decision model, in one forward pass, whether a candidate step is right given the screenshot. It is the most accurate judge we compare on the agent tasks, tied only by Qwen3-VL-32B on Mind2Web, and matches trained process reward models on reasoning; on web and mobile tasks its judgment rests on the screenshot, and one calibrated threshold carries over to unseen websites.
\item \textbf{Where the gains come from.} Across model families, the gain of \method{} over the target grows with how often one of the draft's four candidates is right compared with how often the target is.
\end{itemize}

\section{Related Work}
\label{sec:related}

\paragraph{Speculative Decoding.}
Speculative decoding lets a small model draft several tokens that a large model then checks in one forward pass, accepting them under a rule that leaves the large model's output distribution unchanged~\cite{leviathan2023fast,chen2023accelerating}; EAGLE-3~\cite{li2025eagle3} drafts from the large model's own features and is among the fastest such methods.
For vision-language models, MMSpec~\cite{shen2026mmspec} finds that token-level methods designed for text-only models degrade on multimodal inputs and proposes ViSkip, which adapts speculation to vision tokens.
Spec-VLA~\cite{wang2025specvla} accelerates vision-language-action models by relaxing acceptance according to the distance between action tokens.
Without speculation, vision-language-action models and vision transformers are also accelerated by scheduling models and pruning tokens~\cite{DBLP:conf/iclr/LiMSJTFMXWZ26}, by learning when to think~\cite{DBLP:journals/corr/abs-2605-29438}, and by jointly pruning tokens and channels~\cite{DBLP:journals/pami/LiTMFCMWZ26}, and language models are trained to reason more efficiently~\cite{DBLP:journals/corr/abs-2604-02322}.
These methods work at the level of tokens and keep the large model in every verification step; \method{} works at the level of steps and calls the large model only on the steps it rejects, and the two levels can be combined.

\paragraph{Step-Level Speculative Reasoning.}
RSD~\cite{liao2025rsd} scores each drafted step with a process reward model and has the target model regenerate the step when the score falls below a threshold.
SpecReason~\cite{pan2025specreason} has the target model assess each speculated step and correct it when needed.
ConfSpec~\cite{liu2026confspec} lets the draft model verify its own step and escalates to the target model only when this verification is uncertain, and SpecGuard~\cite{purohit2026tokens} accepts or recomputes a step from attention-based grounding and log-probability signals of the models themselves.
For large multimodal models, DREAM-R~\cite{hu2026dreamr} trains the draft model with reinforcement learning to align its steps with the target and accepts steps through a threshold-based verification rule, and Speculative Verdict~\cite{liu2026verdict} has several small vision-language models draft reasoning paths that a large model combines into the final answer.
Where a pipeline judges each step, \method{} keeps its structure and replaces only that judgment with an external decision model, which is neither the target model, nor a reward model specialised to the task's domain, nor the draft model itself.

\paragraph{Vision-Language Agents.}
Web and mobile benchmarks such as Mind2Web~\cite{deng2023mind2web} and AndroidControl~\cite{li2024androidcontrol} evaluate agents that act on websites and phones from screenshots, and SeeAct~\cite{zheng2024seeact} found that grounding each decision in the screenshot remains the main difficulty for such agents, comparing strategies that include marking candidate elements on the image~\cite{yang2023som}.
Beyond screens, language-model agents are trained with reinforcement learning to use tools~\cite{zheng2026procedureaware} and to act through speech~\cite{DBLP:journals/corr/abs-2608-26432}, and several models reasoning over multiple rounds improve text-to-image generation~\cite{DBLP:journals/corr/abs-2602-06166}.

\paragraph{Speculative Execution in Agents.}
Speculative Actions~\cite{ye2025specactions} uses faster models to predict likely next actions and execute them in parallel, committing only when the prediction matches the agent's own action.
Dynamic Speculative Planning~\cite{guan2026dsp} learns online how far to speculate, trading latency against cost while keeping the result lossless.
DualSpec~\cite{zhong2026dualspec} speculates differently on different action types of deep research agents and checks drafts with a lightweight, confidence-based verifier instead of strict matching, and SpecAgent~\cite{wang2026trajectory} has a weak model draft a segment of tool calls that the strong model reviews in a single prefill pass, committing the longest prefix it judges valid.
These systems check drafts against the large model's own action or its review, so the large model still runs on accepted steps; \method{} judges each drafted step against the screenshot and runs the large model only on the steps it rejects.

\paragraph{Improving the Models by Training.}
Instead of judging the drafts, one can train the models themselves.
Reinforcement learning has improved how agents select behaviours, shape their data distribution and estimate policy gradients, from Atari to continuous control~\cite{DBLP:conf/icml/FanX22,DBLP:conf/iclr/FanZLH0ZWX23,DBLP:journals/corr/abs-2105-03923,DBLP:journals/corr/abs-2106-00707,DBLP:journals/corr/abs-2011-06752,DBLP:journals/corr/abs-2112-04145}, and reward-based fine-tuning carries it to generative models such as flow matching models and protein generators~\cite{DBLP:conf/iclr/FanSCCLL25,DBLP:conf/nips/FanWCCL25,DBLP:journals/corr/abs-2510-18072,DBLP:journals/tmlr/WangFGNJL26}.
For language models, recent work asks whether scaling reinforcement learning requires more training~\cite{yang2026doesscalingreinforcementlearning}, improves the reasoning of audio language models with process rewards and recursive self-improvement~\cite{DBLP:conf/iclr/FanRLPSBGLG26,fan2026audiorsirecursivelyselfimprovingaudio}, and performs on-policy distillation without training~\cite{yang2026trainingfreeonpolicydistillation}.
DREAM-R~\cite{hu2026dreamr} likewise trains its draft model with reinforcement learning; \method{} trains neither model, so it applies unchanged to models improved in any of these ways.

\paragraph{Verifiers, Calibration and Cascades.}
Process reward models score intermediate steps and have been trained from human labels~\cite{lightman2023verify} and from automatically constructed labels~\cite{wang2023mathshepherd}, typically for a single domain; for multimodal reasoning, VisualPRM~\cite{wang2025visualprm} and VL-PRM~\cite{ong2025vlprm} train them on image-grounded steps, and GenRM~\cite{zhang2025genrm} casts verification as next-token prediction, taking the likelihood that the verifier answers ``Yes'' as the score, the form our identically prompted baselines take.
ProcessBench~\cite{zheng2025processbench} evaluates such judges by whether they locate the first erroneous step of a solution.
A threshold on a probability has a fixed meaning only if the probability is calibrated~\cite{guo2017calibration}, and language models can, to a degree, estimate the probability that their own answers are correct~\cite{kadavath2022know}, and the middle layers of an audio language model hold a workspace that can be read out in words~\cite{DBLP:journals/corr/abs-2608-24958}.
Routers decide before generation which model should answer a query~\cite{ong2024routellm}, and cascades judge the full answer of a cheaper model and escalate the query when it falls short~\cite{chen2023frugalgpt}; both decide once per query, whereas \method{} decides once per step, after seeing the drafted step and the image.

\section{Method}
\label{sec:method}

\subsection{Problem Formulation}
\label{sec:setup}

A task is given by an input $x$, which contains the goal or question and an image: a screenshot for an agent, a figure for visual reasoning.
The solution is produced in steps $y_1, y_2, \dots$.
In an agent, each step consists of the short reasoning and the single action it leads to; for mathematical reasoning, it is one segment of the reasoning chain.
We write $c_t = (x, y_{<t})$ for the context of step $t$, which for an agent includes the current screenshot.

Two models are available: a small draft model $q$ and a large target model $p$.
At each step, the draft model proposes $K$ candidate steps $\hat{y}^{(1)}_t, \dots, \hat{y}^{(K)}_t \sim q(\cdot \mid c_t)$, and a judge either selects one of them or rejects them all:
\begin{equation}
y_t =
\begin{cases}
\hat{y}^{(k^\star)}_t & \text{if the judge accepts candidate } k^\star,\\
\tilde{y}_t \sim p(\cdot \mid c_t) & \text{if it rejects all candidates.}
\end{cases}
\label{eq:spec}
\end{equation}
With $K = 1$ this is the single-draft scheme of RSD~\cite{liao2025rsd} and SpecReason~\cite{pan2025specreason}; with $K > 1$ it follows guided speculative inference~\cite{geuter2026gsi}, which scores several draft candidates with a reward model and falls back to the large model when none is good enough.
We call a step correct if the facts it reads from the input and the action or computation it performs are right, and it agrees with the goal and the previous steps.

Existing pipelines differ mainly in the judge (\cref{sec:related}).
A grading judge accepts a draft when a score that $p$ assigns to it exceeds a threshold, so $p$ reads every draft, although it writes only the rejected steps.
A reward-model judge replaces this score with the output of a domain-specific process reward model, and a self-assessment judge uses a signal of $q$ itself, such as its token log-probabilities or its answer to a check question.

\subsection{\method}
\label{sec:vanguard}

\method{} replaces the judge and leaves the rest of \cref{eq:spec} unchanged.
It uses a decision model $D$: a model that takes a context, which may include images, together with a question, and returns the probability of each answer; according to its vendor, it does so in a single forward pass, without generating text.
For each candidate, \method{} forms a query from the input, the image and the candidate as the latest step, and asks a yes-or-no check question $Q$ about whether the latest step is right.
Let
\begin{equation}
r^{(k)}_t = P_D\bigl(\text{yes} \mid x, y_{<t}, \hat{y}^{(k)}_t, Q\bigr)
\end{equation}
be the probability that $D$ assigns to ``yes'' for candidate $k$, and let $k^\star = \arg\max_k r^{(k)}_t$.
\method{} keeps the best candidate when its probability clears a threshold $\tau$, and otherwise hands the step to the target model:
\begin{equation}
\text{accept } \hat{y}^{(k^\star)}_t \iff r^{(k^\star)}_t \ge \tau .
\label{eq:accept}
\end{equation}
The threshold $\tau$ sets the trade-off between cost and accuracy: a higher $\tau$ sends more steps to the target model.
For an agent, candidates that parse to the same action are scored once; \cref{alg:vanguard} summarises one step.

\begin{algorithm}[t]
\caption{One step of speculation with \method{}}
\label{alg:vanguard}
\small
\begin{algorithmic}[1]
\Require context $c_t$ (goal, image, steps so far), draft $q$, target $p$, decision model $D$, question $Q$, $K$, $\tau$
\State Sample candidates $\hat{y}^{(1)}_t,\dots,\hat{y}^{(K)}_t \sim q(\cdot \mid c_t)$
\For{each distinct candidate $\hat{y}^{(k)}_t$}
  \State $r^{(k)}_t \gets P_D(\text{yes} \mid c_t, \hat{y}^{(k)}_t, Q)$ \Comment{one pass, with image}
\EndFor
\State $k^\star \gets \arg\max_k r^{(k)}_t$
\If{$r^{(k^\star)}_t \ge \tau$}
  \State $y_t \gets \hat{y}^{(k^\star)}_t$ \Comment{accept the draft}
\Else
  \State $y_t \gets$ greedy decoding of $p(\cdot \mid c_t)$ \Comment{fall back}
\EndIf
\State \Return $y_t$
\end{algorithmic}
\end{algorithm}

Each check question asks $D$ to check the step against the input rather than to judge its plausibility, and the image is always passed to $D$ together with the text.
For agents, $Q$ is applied to the action of the candidate step; on Mind2Web it reads ``Is this the correct next action toward completing the task? Check that the element is the right one on the screenshot and that any typed or selected value is right.''
For visual reasoning we use the \emph{visual} question, ``Check every fact the step reads from the figure and every computation in it yourself. Are all of them exactly right?''
For text reasoning inside the speculative pipeline we use the \emph{compute} question, ``Check every computation in the latest step yourself. Are all of them exactly right?''; to locate errors in fixed solutions we also use the \emph{correct} question, ``Is the latest step correct? Check its computations, its reasoning, and that it is consistent with the problem and the previous steps.''
$D$ is a model other than $q$, so a draft is never judged by the model that wrote it.

We use Jev~\cite{typesafe2026jev} (release jev-1.13.0) as $D$, queried unchanged through its vendor's public interface; we fit nothing in it, and labels from our tasks are used only to choose the question wording and the threshold on development data.
The cost of $D$ enters every measurement as the wall-clock time of a call, network latency included.

\subsection{Reading the Threshold}
\label{sec:calibration}

Suppose $D$ is calibrated on the candidates it selects: among selected candidates scored $r$, a fraction $r$ is correct.
Then an accepted step is wrong with probability
\begin{equation}
\mathbb{E}\bigl[\, 1 - r^{(k^\star)}_t \mid r^{(k^\star)}_t \ge \tau \,\bigr] \le 1 - \tau ,
\label{eq:bound}
\end{equation}
so $1-\tau$ reads as a tolerated error rate per accepted step.
Because taking the maximum of $K$ scores favours candidates that are scored too high, calibration has to hold after this selection; if $\bigl|P(\text{correct} \mid r^{(k^\star)}_t = r) - r\bigr| \le \epsilon$ for every $r \ge \tau$, the bound becomes $1 - \tau + \epsilon$.
A calibrated judge has a second, practical advantage: the share of steps it accepts at a given $\tau$ depends on how often the drafts are right rather than on the quirks of one dataset, so a threshold chosen on development data keeps both its meaning and its acceptance rate on new data.
We measure calibration and test the transfer of the threshold in \cref{sec:exp_ablation}.

\subsection{When \method{} Saves Time}
\label{sec:cost}

Let $c_K$, $c_D$ and $c_p$ be the average time of drafting the $K$ candidates of one step, of judging the candidates of one step with $D$ and of one target step, and let $\alpha_\tau$ be the fraction of steps accepted at threshold $\tau$, so that $1-\alpha_\tau$ is the share of steps written by the target model.
Assuming that the number of steps does not change, the expected time per step under \method{} is
\begin{equation}
C_{\text{\method}} = c_K + c_D + (1 - \alpha_\tau)\, c_p ,
\end{equation}
which is lower than the time $c_p$ of the target model alone if and only if
\begin{equation}
c_K + c_D < \alpha_\tau\, c_p .
\label{eq:condition}
\end{equation}
Two facts favour agents with reasoning: each step decodes a few sentences of reasoning, so $c_p$ is large, and the $K$ candidates can be sampled in one batch over their shared multimodal context, so $c_K$ grows more slowly than $K$ times the cost of one candidate.
A grading judge that has the target model read each candidate adds $K\,c'_p$ to the left-hand side of \cref{eq:condition}, where $c'_p$ is the time of one such read, while \method{} adds the time of $D$.
Token-level speculative decoding reduces $c_K$ and $c_p$ themselves, so it can be applied inside both terms (\cref{sec:exp_speed}).

\section{Experiments}
\label{sec:exp}

\subsection{Experimental Setup}
\label{sec:exp_setup}

\paragraph{Agent Tasks.}
Multimodal-Mind2Web~\cite{deng2023mind2web,zheng2024seeact}: each step shows a viewport around the gold element with eight candidate elements marked in the style of Set-of-Mark~\cite{yang2023som}, and the agent must output the element, the operation and the value.
We split test\_task by task into development and test parts and use test\_website, whose websites never appear in test\_task, only for testing.
AndroidControl~\cite{li2024androidcontrol}: on three shards of the test set, the agent receives only the high-level goal and must output the next action; a click is correct within a normalised distance of 0.14 of the gold point, and episodes are split likewise.
EB-ALFRED, the household-robot tasks of EmbodiedBench~\cite{yang2025embodiedbench} built on ALFRED~\cite{shridhar2020alfred}: from an egocentric image, the instruction and the executed actions, the agent chooses the next high-level action among eight options, many distractors being infeasible in the current state; episodes are split likewise.
In these three tasks the agent writes two to four sentences of reasoning before each action, every step is evaluated given the gold history, and the metric is step success: the whole action must be correct.
To let the agent's own errors accumulate, we also run complete closed-loop episodes of two BabyAI tasks~\cite{chevalierboisvert2019babyai}, GoToLocal and PickupLoc, with thresholds chosen on development episodes; the room is rendered with row and column labels and the agent moves in absolute directions (\cref{sec:s_closed_loop}), and the metric is episode success.

\paragraph{Models and Judges.}
The draft and target models are Qwen3-VL-2B and Qwen3-VL-8B (Instruct)~\cite{bai2025qwen3vl}, served with vLLM~\cite{kwon2023vllm}; the draft samples $K = 4$ candidates at temperature 0.7, and the target decodes greedily.
We compare \method{} with judges of every kind used in step-level speculation: random choice among the candidates, a vote gate that accepts only when the candidates agree, the draft's mean token log-probability, the draft checking its own candidates, the vision-language process reward model Qwen-VL-PRM-7B~\cite{ong2025vlprm}, the target model grading each candidate from 0 to 9 as in SpecReason~\cite{pan2025specreason}, and the target model and Qwen3-VL-32B asked exactly \method{}'s question, scored by their probability of ``yes''~\cite{zhang2025genrm}.

\begin{figure}[t]
\centering
\includegraphics[width=\linewidth]{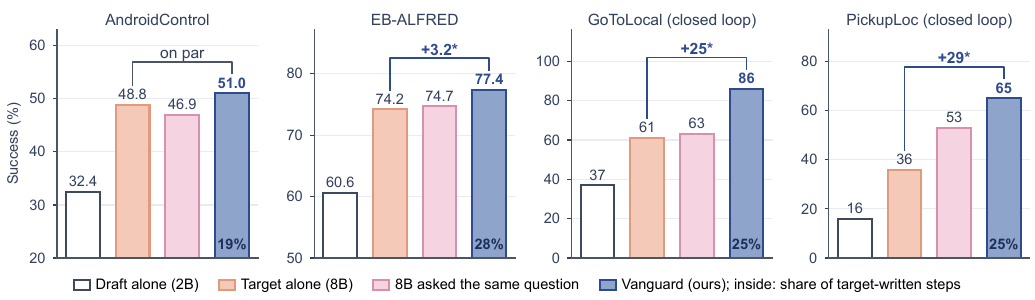}
\caption{\textbf{With \method{}, the 2B draft is at least as accurate as its 8B target on every agent task, and far more accurate in closed loop.} Step success on AndroidControl (525 steps) and EB-ALFRED (1,849); episode success on 100 test episodes per BabyAI task; higher is better. Brackets: comparison with the target alone (*: significant gain); inside: share of target-written steps.}
\label{fig:agents}
\end{figure}

\paragraph{Protocol.}
Thresholds are chosen on development data and fixed on the test sets, except for the AndroidControl baselines of \cref{fig:agents}, shown at their most accurate test point; \method{} uses $\tau = 0.5$ on Mind2Web and AndroidControl and 0.451 on EB-ALFRED unless stated otherwise, and in the agent comparisons of \cref{sec:exp_judges,sec:exp_ablation} every judge's threshold sends about the same share of steps to the target.
Significance is assessed with the exact two-sided McNemar test on paired steps.
Times are measured at batch size 1 on NVIDIA RTX A6000 GPUs, including the latency of the decision model's public API, end to end in \cref{sec:exp_speed} and per component elsewhere (\cref{sec:s_details}).

\paragraph{Implementation Details.}
For the agents, each candidate is a short paragraph of reasoning followed by one action, which is parsed into the element, operation and value, or into the action type and its arguments; candidates that parse to the same action are scored once.
The judge is shown the screenshot, the goal, the previous actions and the proposed action, but not the candidate's reasoning; on Mind2Web it also sees the list of candidate elements, and on AndroidControl the proposed tap is marked on the screenshot with a red circle; the target asked the same question receives the same inputs.
The target writes a rejected step greedily from the same context; for reasoning, each paragraph of at least 40 tokens is one step, and the draft continues from the accepted or rewritten step.

\subsection{Main Results: A Small Draft Model Surpasses the Target}
\label{sec:exp_main}

\begin{table}[t]
\centering
\small
\setlength{\tabcolsep}{1.5pt}
\begin{tabular}{@{}lcccc@{}}
\toprule
& \multicolumn{2}{c}{test\_task} & \multicolumn{2}{c}{test\_website} \\
\cmidrule(lr){2-3}\cmidrule(l){4-5}
Method & Success$\uparrow$ & 8B steps$\downarrow$ & Success$\uparrow$ & 8B steps$\downarrow$ \\
\midrule
Draft model alone (2B) & 42.4 & 0 & 43.8 & 0 \\
Target model alone (8B) & 64.4 & 100 & 63.3 & 100 \\
\midrule
Random of four & 45.0 & 0 & 42.6 & 0 \\
Vote gate & 56.1 & 33 & 56.9 & 37 \\
Qwen-VL-PRM-7B & 63.0 & 36 & 61.0 & 39 \\
8B, same prompt & 65.5 & 23 & 65.1 & 22 \\
\rowcolor{vgmist}\textcolor{vgnavy}{\method{} (ours)} & \textbf{66.5} & 26 & \textbf{66.8} & 23 \\
\bottomrule
\end{tabular}
\caption{\textbf{With \method{}, the 2B draft leads every method in this table on both held-out Mind2Web test sets.} Step success (\%) with four draft candidates and 8B fallback on 820 / 975 held-out steps; 8B steps: share written by the 8B (23\% on development at $\tau = 0.5$); thresholds fixed on development data, each baseline at the share closest to \method{}'s, the PRM at its single development threshold; significance tests: \cref{sec:s_fulltables,sec:s_pairs}.}
\label{tab:m2w}
\end{table}

\begin{figure}[t]
\centering
\includegraphics[width=0.6\linewidth]{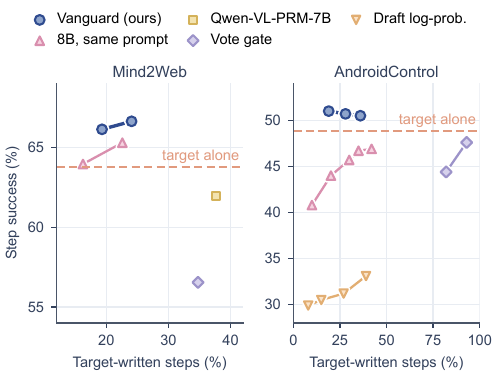}
\caption{\textbf{\method{} lies above the target model alone on Mind2Web and above the target asked the same question on AndroidControl.} Step success against the share of target-written steps on test steps (higher and further left is better): Mind2Web (both test sets, development thresholds) and AndroidControl (development sweep; \method{} from 19\% target-written steps).}
\label{fig:share}
\end{figure}

\begin{figure}[t]
\centering
\includegraphics[width=0.7\linewidth]{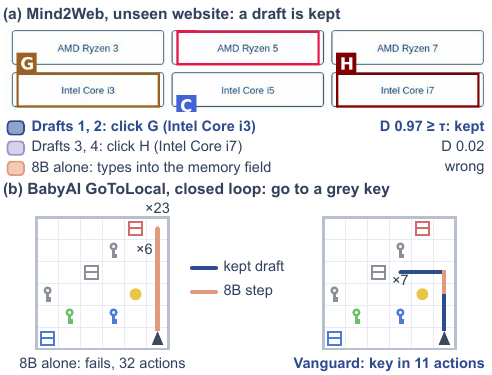}
\caption{\textbf{\method{} keeps a draft when the screen confirms it and otherwise hands the step to the target.} (a) A held-out Mind2Web step, unseen website: $D$ separates the right drafts from the wrong ones and the step is kept; the 8B alone types into a later field. (b) One closed-loop BabyAI episode: navy, steps kept from the 2B draft; peach, steps written by the 8B (for \method{}, after a rejection); $\times n$: actions in that cell; the 8B alone fails after 32 actions, and \method{} reaches the key in 11. Both cases are from Figs.~\labelcref{fig:s_cases_win}a and~\labelcref{fig:s_cases_loop}a.}
\label{fig:case}
\end{figure}

\paragraph{On the web, a 2B draft surpasses its 8B target.}
On both held-out Mind2Web test sets, the 2B draft judged by \method{} is more accurate than the 8B alone, significantly on the unseen websites and on the two sets together, while the target writes only a minority of the steps (\cref{tab:m2w,fig:share}).
Its four drafts often contain a correct action where the target's own action is wrong, and \method{} recognises these drafts (\cref{fig:case}a).
The other judges in \cref{tab:m2w} share the same candidates and fallback, yet none of them surpasses the target significantly, and at a similar share of 8B steps \method{} stays ahead of the target asked the same question.

\paragraph{The draft matches the target on phones and beats it in homes and in closed loop.}
On AndroidControl, \method{} matches the 8B alone with fewer target-written steps and is the best judge with and without fallback, picking among the four candidates without fallback significantly better than the target asked the same question (\cref{sec:s_fulltables}).
On the embodied tasks it is significantly more accurate than the target, whereas the target asked the same question writes most steps itself and still stays behind (\cref{fig:agents}).
In closed loop, where each step builds on the agent's own previous actions, better steps add up: the 2B draft completes far more episodes than the 8B alone (\cref{fig:agents,fig:case}b), while the 8B asked the same question hands most of the decisions back to itself.

\subsection{When Does a Small Draft Beat the Target?}
\label{sec:exp_rule}

\begin{figure}[t]
\centering
\includegraphics[width=0.6\linewidth]{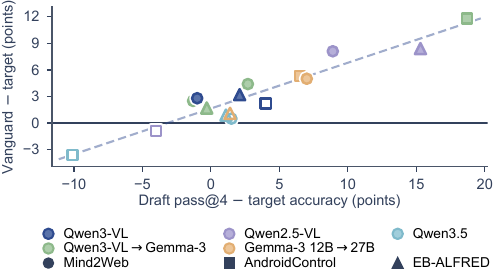}
\caption{\textbf{\method{}'s gain over the target grows with the draft's headroom.} Each point is one draft/target pair on one agent task, all sixteen; $x$: headroom (draft pass@4 minus target accuracy), $y$: \method{} minus target (higher is better); filled: significant gain; dashed: least-squares fit, Pearson $r = 0.95$ (\cref{sec:s_passk}).}
\label{fig:passk}
\end{figure}

\begin{table}[t]
\centering
\footnotesize
\setlength{\tabcolsep}{4pt}
\renewcommand{\arraystretch}{1.12}
\begin{tabular}{@{}llcc@{}}
\toprule
Draft $\rightarrow$ target & Task & Target\,$\uparrow$ & \method{}\,$\uparrow$ \\
\midrule
Qwen2.5-VL 3B $\rightarrow$ 7B & Mind2Web & 50.5 & \textbf{58.6} \\
& EB-ALFRED & 67.9 & \textbf{76.3} \\
\addlinespace[3pt]
Qwen3-VL-2B $\rightarrow$ & Mind2Web & 64.1 & \textbf{66.6} \\
\quad Gemma-3-27B & AndroidControl & 34.1 & \textbf{45.9} \\
& EB-ALFRED & 76.6 & \textbf{78.3} \\
\addlinespace[3pt]
Gemma-3 12B $\rightarrow$ 27B & Mind2Web & 64.1 & \textbf{69.1} \\
& AndroidControl & 34.1 & \textbf{39.4} \\
\bottomrule
\end{tabular}
\caption{\textbf{In these further pairs, also across families, \method{} lifts the draft significantly above the target.} Step success (\%) on the test steps (Mind2Web: both test sets), at each pair's development threshold; every gain is significant (largest $p = 0.025$; tests and further pairs: \cref{sec:s_families,sec:s_cross}).}
\label{tab:beyond}
\end{table}

\paragraph{The gain grows with the draft's headroom.}
A judge can only choose among its candidates, so a draft can surpass its target only where its candidates are right and the target is not.
Across four model families (Qwen3-VL, Qwen2.5-VL~\cite{bai2025qwen25vl}, Qwen3.5~\cite{qwen2026qwen35}, Gemma-3~\cite{gemmateam2025gemma3}) and an unchanged protocol, the gain of \method{} grows with the draft's \emph{headroom}, its pass@4 minus the target's accuracy (\cref{fig:passk}); where the headroom is largest, \method{} turns more than half of it into a significant gain (\cref{tab:beyond}).

\paragraph{The gain is not tied to one model family.}
Step-level speculation exchanges whole steps, not tokens, so the target may come from another family: the Qwen3-VL-2B draft significantly surpasses Gemma-3-27B on all three agent tasks, even where its headroom is close to zero, and a Gemma-3-12B draft, whose headroom over Gemma-3-27B we measured on development steps before running the judge, significantly surpasses the 27B on Mind2Web and AndroidControl (\cref{tab:beyond}).

\begin{figure}[t]
\centering
\includegraphics[width=0.6\linewidth]{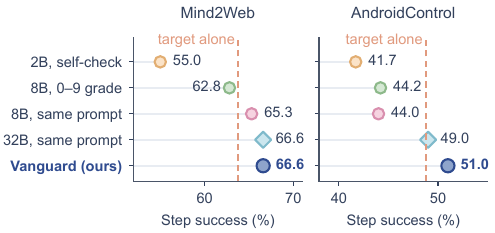}
\caption{\textbf{\method{} is ahead of every other judge on both tasks, except for a tie with the 32B asked the same question on Mind2Web.} Higher is better. Every judge scores the same candidates at a development threshold sending about the same share of steps to the 8B (full table: \cref{tab:judges}).}
\label{fig:judges}
\end{figure}

\begin{figure}[t]
\centering
\includegraphics[width=\linewidth]{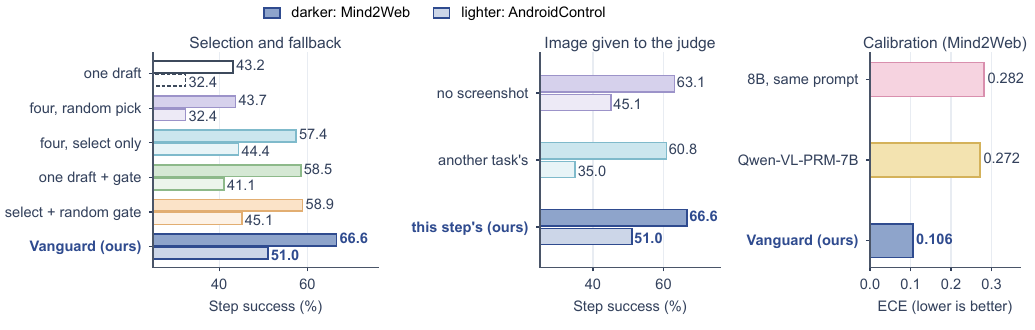}
\caption{\textbf{Selection and fallback are both needed, the judge must see this step's screenshot, and its probabilities are the best calibrated.} Test steps; higher step success and lower calibration error are better; gated and image variants at about the same share of target-written steps as \method{}; right: calibration error on Mind2Web. Full tables: \cref{tab:ablation,tab:screenshot,tab:ece}.}
\label{fig:ablation}
\end{figure}

\subsection{Comparison with Other Judges}
\label{sec:exp_judges}

\paragraph{\method{} is the most accurate judge we compare.}
The gap is largest against the draft checking its own candidates (\cref{fig:judges}), and grading on a ten-point scale, the scheme of SpecReason, is significantly worse on both tasks although it uses the target model itself.
The target asked the same question is close on Mind2Web but falls significantly behind on AndroidControl, and Qwen3-VL-32B asked the same question comes closest, tying on Mind2Web.
Across the nine (family, task) cells of the three Qwen families, \method{} is significantly more accurate than the vision-language process reward model Qwen-VL-PRM-7B in all nine and than the target asked the same question in seven; it is also significantly more accurate than GUI-Critic-R1~\cite{wanyan2025guicritic}, a model trained to judge GUI actions, on both tasks, and, with a Gemma-3-27B target, than that target asked the same question on all three tasks (\cref{sec:s_cross,sec:s_judges_all}).

\paragraph{\method{}'s scores, unlike the target's, are graded.}
Asked the same question, the 8B puts almost all of its probability near zero or one, so a threshold on it accepts or rejects nearly everything, whereas \method{} spreads its scores and separates right from wrong candidates over the whole range (\cref{sec:s_where}).

\paragraph{The same judge carries over to mathematical reasoning.}
With only the check question changed (\cref{sec:vanguard}), on MathVista~\cite{lu2024mathvista} with the agent models and on ProcessBench~\cite{zheng2025processbench} with Qwen3~\cite{yang2025qwen3}, \method{} is on par with process reward models trained for these domains~\cite{ong2025vlprm,zhang2025lessons} and ahead of general models asked the same question, significantly so against the 8B on MathVista at the lower share of target-written steps, and picking one of four solutions on MATH-500~\cite{hendrycks2021math,lightman2023verify} it significantly beats the math process reward model (\cref{sec:s_reasoning}).

\subsection{Ablation Study}
\label{sec:exp_ablation}

\begin{figure}[t]
\centering
\includegraphics[width=0.6\linewidth]{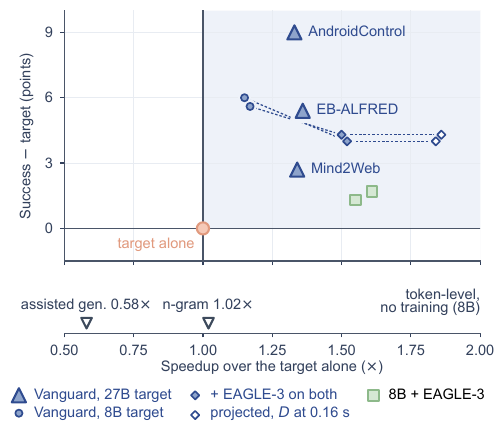}
\caption{\textbf{\method{} is faster and more accurate than its target; without EAGLE-3, it saves the most time with a 27B target from another family.} Shaded: faster and more accurate than the target. Same steps per method and target (8B: 300 Mind2Web steps per test set, $\tau = 0.41$ on test\_task; Gemma-3-27B: 300, 200 and 300 steps of Mind2Web, AndroidControl and EB-ALFRED, \method{} 67.7, 45.0 and 83.7\% against 65.0, 36.0 and 78.3\%); hollow diamonds: $D$ at its nominal latency; bottom: token-level methods without training (50 and 200 Mind2Web steps).}
\label{fig:speed}
\end{figure}

\paragraph{Selection and fallback are both needed.}
In \cref{fig:ablation} (left), choosing among four candidates without fallback already lifts the draft model far above random choice, so the decision model recognises good actions; still, a step that all four candidates get wrong cannot be repaired, and selection alone stays below the target model, as do a gate on a single draft and \method{}'s selection with a gate that falls back at random.
Only the combination significantly surpasses the target model on Mind2Web: \method{} accepts a candidate exactly when it is confident, and on the Mind2Web steps it accepts, its chosen action is right more often than the target model's own action, while on the steps it rejects its best candidate is usually wrong (\cref{sec:s_where}).
More candidates help too: as $K$ grows from one to four, accuracy rises significantly on Mind2Web, where the share of target-written steps also falls (\cref{sec:s_seeds,sec:s_pairs}).

\paragraph{The judge must see the screenshot.}
Rescoring the same candidates with only the image changed (\cref{fig:ablation}, middle), on both tasks \method{} falls to or below the level of the target model without the screenshot, and with a screenshot from another task it falls significantly below it.
The gate's ability to separate right from wrong candidates drops accordingly, so the judgment rests on comparing the action with the screen; on MathVista, removing the figure costs significant accuracy as well (\cref{sec:s_inputs}).

\paragraph{Calibrated scores let one threshold transfer.}

Over the selected candidates on Mind2Web (\cref{fig:ablation}, right), \method{}'s probabilities are far better calibrated than those of the target asked the same question or of the vision-language process reward model.
As \cref{sec:calibration} predicts, at $\tau = 0.5$ the share of target-written steps barely moves from development data to held-out tasks and unseen websites (\cref{tab:m2w}).

\subsection{Speed and Token-Level Speculation}
\label{sec:exp_speed}

\paragraph{With an expensive target from another family, \method{} is both faster and more accurate.}
Every fallback to a 27B target is costly: with the 2B draft, \method{} is faster than the 27B alone on all three agent tasks (\cref{fig:speed}) and significantly more accurate on full test sets (\cref{tab:beyond}).

\paragraph{The saving is predictable, and the accuracy gain adds to token-level speculation.}
With the 8B target, drafting four candidates and judging them takes less time than the target steps that \method{} saves on Mind2Web and AndroidControl, so \cref{eq:condition} holds on both, and these times with the fallback rate predict the measured speedup in all nine timed 8B settings (\cref{sec:s_cost}).
With EAGLE-3~\cite{li2025eagle3} inside both models, \method{} reaches nearly the speed of EAGLE-3 alone while staying more accurate on Mind2Web, whereas training-free token-level methods give the 8B little or no speedup (\cref{fig:speed}).

\section{Conclusion}
\label{sec:conclusion}

A few samples of a small model can contain the right step about as often as a large model finds it; with \method{}, the small draft is at least as accurate as its target on web, mobile and embodied agents and significantly more accurate on Mind2Web and EB-ALFRED, and with a 27B target from another family it is both faster and more accurate; in closed loop it completes more episodes.
\method{} is the most accurate judge we compare on the agent tasks, and its gain over the target grows with the draft's pass@4 relative to the target's accuracy.
Step-level speculation has so far been measured by how little accuracy it gives up; needing neither training nor the target's weights, a judge that checks each step against the screen lets it raise accuracy instead.

\clearpage
\bibliographystyle{unsrtcase}
\bibliography{main}

\clearpage
\appendix
\crefalias{section}{appendix}\crefalias{subsection}{subappendix}\crefalias{subsubsection}{subsubappendix}

\section{Discussion}
\label{sec:s_discussion}

\subsection{The Problem \method{} Addresses}
\label{sec:s_disc_problem}
Vision-language agents browse websites, operate phones and act in households by reading a screenshot or a camera image, reasoning about the goal and choosing an action, one step after another.
The agents that work best reason at length before every action, so each step is a full decoding pass of a large model, and the better the agent, the more every step costs.
Step-level speculation lets a small draft model write the steps and calls the large target model only on the steps a judge rejects.
Everything then depends on the judge: it decides what is saved, what is lost, and whether the small model can end up ahead of the large one.
\method{} is a training-free judge for this setting that checks each drafted step against the screen.

\clearpage
\subsection{Why Existing Speculation Falls Short for Visual Agents}
\label{sec:s_disc_gap}
Token-level speculative decoding checks the draft against the large model's own distribution; this keeps the output exact but has the large model score every token, needs draft heads trained for the target or a tokenizer shared with it, and its methods built for text-only models degrade on multimodal inputs.
Step-level methods relax this requirement, and their judges fall into three groups.
A grading judge asks the large model to grade each draft, so the large model still reads every drafted step, and its probabilities pile up near zero and one.
A process reward model is specialised to one domain and must be trained for it.
A self-assessment judge relies on signals of the small model itself and inherits whatever the small model fails to notice about its own mistakes.
Visual inputs add a further difficulty: a step can be fluent and logically sound, yet click an element that is not on the screen or read the wrong value off a chart, and a judge that sees only the text of a step cannot tell whether the step agrees with the image.

\clearpage
\subsection{Knowing When to Trust the Draft}
\label{sec:s_disc_idea}
Sampled a few times, a small vision-language model can propose a correct step about as often as the large model answers correctly, and not always on the same steps.
\method{} turns this into accuracy by asking, for each candidate, a general decision model a yes-or-no question about whether the step is right, with the task and the image in view, and reading off the probability of ``yes'' in a single forward pass.
The question asks the decision model to check the step against the input rather than to judge its plausibility, and the decision model is a model other than the draft, so a draft is never judged by the model that wrote it.
The highest-scoring candidate is kept when its probability clears a threshold $\tau$, and otherwise the target model writes the step.
When the decision model is calibrated on the candidates it selects, $1-\tau$ reads as a tolerated error rate per accepted step, and the share of accepted steps depends on how often the drafts are right rather than on one dataset, so a threshold chosen on development data keeps its meaning on new data.
Nothing is trained: only the wording of the question and the threshold are chosen on development data, and the draft and target models are left unchanged, so the target may come from another model family.

\clearpage
\subsection{What It Costs}
\label{sec:s_disc_cost}
With $c_K$, $c_D$ and $c_p$ the average times of drafting the $K$ candidates of one step, of judging them with $D$ and of one target step, and $\alpha_\tau$ the share of accepted steps, \method{} is faster than the target model alone, for an unchanged number of steps, if and only if $c_K + c_D < \alpha_\tau\, c_p$ (\cref{eq:condition} in \cref{sec:cost}).
Agents with reasoning favour this condition: each target step decodes a few sentences of reasoning, so $c_p$ is large, and the $K$ candidates are sampled in one batch over their shared multimodal context.
Where a grading judge adds $K$ reads of the target model to every step, \method{} adds the time $c_D$ of judging the candidates with $D$, one forward pass per distinct candidate, whose wall-clock time, network latency included, enters every measured speedup.
Token-level speculative decoding reduces the drafting and target terms themselves, so the two levels of speculation compose.

\clearpage
\section{Algorithm Pseudocode}
\label{sec:s_algorithm}

\Cref{alg:s_episode} gives the full speculative loop of \method{} over the steps of one task, in the notation of the main text: input $x$, steps $y_1, y_2, \dots$, context $c_t = (x, y_{<t})$, draft model $q$, target model $p$, decision model $D$, check question $Q$, number of candidates $K$ and threshold $\tau$.
For an agent, $c_t$ includes the current screenshot, and candidates that parse to the same action are scored once.

\begin{algorithm}[!htb]
\caption{Speculation with \method{} over one task}
\label{alg:s_episode}
\begin{algorithmic}[1]
\Require input $x$ (goal or question, and image), draft model $q$, target model $p$, decision model $D$, check question $Q$, number of candidates $K$, threshold $\tau$
\State $t \gets 1$
\While{the task is not finished}
  \State $c_t \gets (x, y_{<t})$ \Comment{for an agent, with the current screenshot}
  \State Sample candidates $\hat{y}^{(1)}_t,\dots,\hat{y}^{(K)}_t \sim q(\cdot \mid c_t)$ \Comment{in one batch}
  \For{each distinct candidate $\hat{y}^{(k)}_t$}
    \State $r^{(k)}_t \gets P_D\bigl(\text{yes} \mid x, y_{<t}, \hat{y}^{(k)}_t, Q\bigr)$ \Comment{one forward pass, with the image}
  \EndFor
  \State $k^\star \gets \arg\max_k r^{(k)}_t$
  \If{$r^{(k^\star)}_t \ge \tau$}
    \State $y_t \gets \hat{y}^{(k^\star)}_t$ \Comment{accept the draft}
  \Else
    \State $y_t \gets$ greedy decoding of $p(\cdot \mid c_t)$ \Comment{fall back to the target model}
  \EndIf
  \State Execute or append $y_t$; \ $t \gets t + 1$
\EndWhile
\State \Return $y_1, y_2, \dots$
\end{algorithmic}
\end{algorithm}
\clearpage
\section{Experimental Details}
\label{sec:s_details}

\paragraph{Data and Splits.}
Multimodal-Mind2Web: we keep the steps whose gold element has a bounding box and whose viewport contains at least seven other distinct candidate elements (1,255 of the 1,339 steps of test\_task and 975 of the 1,019 of test\_website); the test\_task steps are hashed by task into a development part of 435 steps and a test part of 820 steps, and test\_website is used only for testing.
AndroidControl: three shards of the test set (728 steps), hashed by episode into 203 development and 525 test steps.
MathVista: the 1,000 testmini problems are split at random (seed 0) into 200 development and 800 test problems.
MATH-500: 100 development and 400 test problems.
ProcessBench: 3,400 solutions (GSM8K 400, MATH 1,000, OlympiadBench 1,000, Omni-MATH 1,000); thresholds are chosen on GSM8K.

\paragraph{Thresholds.}
For every judge, candidate thresholds are quantiles of its scores on the development part at several shares of target-written steps, plus 0.5; they are then fixed and applied unchanged to the test sets, except for the AndroidControl baselines of \cref{fig:agents} and \cref{tab:android}, shown at their most accurate test point.
\method{} uses $\tau = 0.5$ on Mind2Web and AndroidControl, $0.451$ on EB-ALFRED and $0.3$ on BabyAI unless stated otherwise; the full sweeps on Mind2Web and AndroidControl are in \cref{sec:s_agents}.

\paragraph{Timing.}
Unless stated otherwise, all times are measured at batch size 1, one request after another, on NVIDIA RTX A6000 GPUs with vLLM~0.19, each model on its own GPU, and with the code, weights and outputs on local disk.
In the per-component timings, single measurements above 5\,s (file-system stalls) are replaced by the median of that component.
We query the decision model over its public API; its latency (0.4--0.6\,s per step during the agent runs, against a nominal 0.16\,s) is included in every measured speedup; we also report speedups at the nominal latency.

\clearpage
\section{Additional Results}
\label{sec:s_results}
\suppressfloats[t]

\subsection{Full Tables of the Main-Text Figures}
\label{sec:s_fulltables}
This section gives the full numbers behind the figures of the main text, with their significance tests: \Cref{fig:agents} corresponds to \cref{tab:android,tab:embodied}, \cref{fig:judges} to \cref{tab:judges}, \cref{fig:ablation} to \cref{tab:ablation,tab:screenshot,tab:ece} and \cref{fig:speed} to \cref{tab:speed}; \Cref{fig:share,fig:passk} are covered in \cref{sec:s_agents,sec:s_passk}, and the rows of \cref{tab:beyond} come from \cref{tab:s_families,tab:s_cross} (Gemma-3 12B$\to$27B: \cref{sec:s_cross}); the Gemma-3-27B points and token-level baselines of \cref{fig:speed} are in \cref{sec:s_cross,sec:s_compose_full,sec:s_short_drafts}, and \cref{sec:s_pairs} collects further paired tests.

\begin{table}[!htb]
\centering
\small
\setlength{\tabcolsep}{4pt}
\begin{tabular}{@{}lcc@{}}
\toprule
Judge & No fallback & With fallback \\
\midrule
Draft log-probability & 29.7 & 33.1 (39) \\
Vote gate & 30.1 & 47.6 (93) \\
Qwen3-VL-8B, same prompt & 39.6 & 46.9 (42) \\
\rowcolor{vgmist}\method{} & \textbf{44.4} & \textbf{51.0} (19) \\
\midrule
\multicolumn{3}{@{}l}{Draft alone 32.4 \quad Random of four 32.4 \quad Target alone 48.8} \\
\bottomrule
\end{tabular}
\caption{\textbf{On AndroidControl, \method{} is the most accurate judge, with and without fallback.} AndroidControl step success (\%) on 525 test steps with four candidates per step, without fallback (the judge must pick a candidate) and with 8B fallback; the share of target-written steps (\%) is in parentheses. \method{} uses its development threshold; each baseline is shown at its most accurate test point, which favours the baselines. Without fallback, \method{} beats the 8B asked the same question on 35 steps against 10 ($p = 0.0002$).}
\label{tab:android}
\end{table}
\FloatBarrier

On AndroidControl (\cref{tab:android}), \method{} is 2.2 points above the 8B alone (51.0 against 48.8\%) and stays above it in all three draws of the draft's samples (\cref{tab:s_seeds}) and at every threshold that sends 19--36\% of the steps to the 8B (\cref{tab:s_android}); it reaches this level while the 8B writes only 19\% of the steps, whereas the most accurate baseline with fallback in \cref{tab:android}, the vote gate, needs 93\% of them to reach 47.6\%.
\FloatBarrier

\begin{table}[!htb]
\centering
\small
\setlength{\tabcolsep}{3.2pt}
\begin{tabular}{@{}lcccccc@{}}
\toprule
& \multicolumn{2}{c}{EB-ALFRED} & \multicolumn{2}{c}{GoToLocal} & \multicolumn{2}{c}{PickupLoc} \\
\cmidrule(lr){2-3}\cmidrule(lr){4-5}\cmidrule(l){6-7}
Method & Succ. & Target & Succ. & Target & Succ. & Target \\
\midrule
Draft alone (2B) & 60.6 & 0 & 37 & 0 & 16 & 0 \\
Target alone (8B) & 74.2 & 100 & 61 & 100 & 36 & 100 \\
8B, same prompt & 74.7 & 66 & 63 & 90 & 53 & 83 \\
\rowcolor{vgmist}\method{} & \textbf{77.4} & 28 & \textbf{86} & 25 & \textbf{65} & 25 \\
\bottomrule
\end{tabular}
\caption{\textbf{On the embodied tasks, \method{} surpasses the 8B significantly while the 8B writes about a quarter of the steps or decisions.} EB-ALFRED step success (\%) on 1,849 test steps, and closed-loop episode success (\%) on 100 test episodes of two BabyAI tasks; ``Target'' is the share (\%) of steps or decisions written by the 8B, all thresholds are chosen on development data (\method{}: 0.451 on EB-ALFRED, 0.3 on BabyAI). \method{} against the 8B alone: 132/73 steps on EB-ALFRED ($p = 5\times10^{-5}$), 27/2 episodes on GoToLocal and 34/5 on PickupLoc ($p = 2\times10^{-6}$ for both).}
\label{tab:embodied}
\end{table}
\FloatBarrier

On EB-ALFRED and in both closed-loop BabyAI tasks, \method{} is the most accurate row of \cref{tab:embodied}, while the 8B writes about a quarter of the steps or decisions.
\FloatBarrier

\begin{table}[!htb]
\centering
\small
\setlength{\tabcolsep}{4pt}
\begin{tabular}{@{}lcc@{}}
\toprule
Judge & Mind2Web & AndroidControl \\
\midrule
Draft model alone (2B) & 43.2 & 32.4 \\
Target model alone (8B) & 63.8 & 48.8 \\
\midrule
Qwen3-VL-2B, self-check & 55.0 & 41.7 \\
Qwen3-VL-8B, 0--9 grade & 62.8 & 44.2 \\
Qwen3-VL-8B, same prompt & 65.3 & 44.0 \\
Qwen3-VL-32B, same prompt & \textbf{66.6} & 49.0 \\
\rowcolor{vgmist}\method{} & \textbf{66.6} & \textbf{51.0} \\
\bottomrule
\end{tabular}
\caption{\textbf{\method{} is ahead of every other judge on both tasks, except for a tie with the 32B asked the same question on Mind2Web.} Step success (\%) of the same pipeline with different judges, each scoring the same candidates, with thresholds chosen on development data so that every judge sends about the same share of steps to the 8B as \method{} (Mind2Web: both test sets, 1,795 steps; AndroidControl: 525 test steps). Against \method{}, the 2B self-check wins/loses 84/293 steps and the 0--9 grade 69/137 on Mind2Web ($p < 10^{-5}$ for both), and the 8B asked the same question wins/loses 14/51 on AndroidControl ($p = 4\times10^{-6}$).}
\label{tab:judges}
\end{table}
\FloatBarrier

At a similar share, \method{} is significantly more accurate than the 2B self-check and the 0--9 grade on both tasks and than the 8B asked the same question on AndroidControl (\cref{tab:judges,tab:s_pairs}); the 32B asked the same question comes closest (49.0\% on AndroidControl), running a 32B model on every candidate.
\FloatBarrier

\begin{table}[!htb]
\centering
\small
\setlength{\tabcolsep}{3.5pt}
\begin{tabular}{@{}ccccc@{}}
\toprule
$K$ & Selection & Gate & Mind2Web & AndroidControl \\
\midrule
1 & -- & $\times$ & 43.2 & 32.4 \\
4 & random & $\times$ & 43.7 & 32.4 \\
4 & \checkmark & $\times$ & 57.4 & 44.4 \\
1 & -- & \checkmark & 58.5 & 41.1 \\
4 & \checkmark & random & 58.9 & 45.1 \\
\rowcolor{vgmist}4 & \checkmark & \checkmark & \textbf{66.6} & \textbf{51.0} \\
\midrule
\multicolumn{3}{@{}l}{Target model alone (8B)} & 63.8 & 48.8 \\
\bottomrule
\end{tabular}
\caption{\textbf{Only selection and fallback together lift the draft above the target.} Step success (\%) on the test steps. ``Selection'' picks the candidate with the highest probability from $D$; the fallback gate hands the step to the 8B when that probability is below $\tau$. Rows with a gate send the same share of steps to the 8B as \method{}.}
\label{tab:ablation}
\end{table}
\FloatBarrier

Selection alone, a gate alone and selection with a random gate all stay below the 8B alone; only selection and fallback together, as in \method{}, lift the draft above it on both tasks (\cref{tab:ablation}).
\FloatBarrier

\begin{table}[!htb]
\centering
\small
\setlength{\tabcolsep}{3pt}
\begin{tabular}{@{}lcccc@{}}
\toprule
& \multicolumn{2}{c}{Mind2Web} & \multicolumn{2}{c}{AndroidControl} \\
\cmidrule(lr){2-3}\cmidrule(l){4-5}
Image given to the judge & Success $\uparrow$ & AUROC $\uparrow$ & Success $\uparrow$ & AUROC $\uparrow$ \\
\midrule
No screenshot & 63.1 & 0.764 & 45.1 & 0.529 \\
Another task's screenshot & 60.8 & 0.673 & 35.0 & 0.544 \\
\rowcolor{vgmist}This step's screenshot & \textbf{66.6} & \textbf{0.807} & \textbf{51.0} & \textbf{0.714} \\
\bottomrule
\end{tabular}
\caption{\textbf{The judge must see the screenshot.} Step success (\%) and AUROC of the gate on the test steps. Each variant rescores the same candidates, with its threshold chosen on development data to send about the same share of steps to the 8B. The correct screenshot beats no screenshot on 108 steps against 45 on Mind2Web ($p = 4\times10^{-7}$) and on 57 against 26 on AndroidControl ($p = 0.0009$); with another task's screenshot, the pipeline falls below the 8B alone, which wins 207 steps against 154 on Mind2Web ($p = 0.006$) and 99 against 27 on AndroidControl ($p = 7\times10^{-11}$).}
\label{tab:screenshot}
\end{table}
\FloatBarrier

With no screenshot or with another task's screenshot, step success drops on both tasks, so the judgment of $D$ rests on the screen it is shown (\cref{tab:screenshot}).
\FloatBarrier

\begin{table}[!htb]
\centering
\small
\setlength{\tabcolsep}{6pt}
\begin{tabular}{@{}lc@{}}
\toprule
Judge & Expected calibration error $\downarrow$ \\
\midrule
Qwen3-VL-8B, same prompt & 0.282 \\
Qwen-VL-PRM-7B & 0.272 \\
\rowcolor{vgmist}\method{} & \textbf{0.106} \\
\bottomrule
\end{tabular}
\caption{\textbf{\method{}'s scores are far better calibrated.} Expected calibration error of the selected candidate's score on Mind2Web (ten bins).}
\label{tab:ece}
\end{table}
\FloatBarrier

At the shares of \cref{tab:m2w}, \method{} also stays ahead of the 8B asked the same question (66.5 and 66.8\% against 65.5 and 65.1\%, 66.6 against 65.3\% pooled), and its scores are far better calibrated (\cref{tab:ece}).
\FloatBarrier

\FloatBarrier

For the main Mind2Web table (\cref{tab:m2w}), \method{} against the 8B alone wins 61 and loses 44 steps on test\_task, wins 95 and loses 61 on test\_website ($p = 0.008$), and wins 156 and loses 105 pooled ($p = 0.002$); against the 8B asked the same question at the same threshold $\tau = 0.5$, it wins 108 and loses 60 pooled ($p = 0.0003$).
On test\_task at $\tau = 0.41$, \method{} is above the 8B in each of the three draws of the draft's samples (66.5, 66.3 and 66.6\% against 64.4\%; \cref{tab:s_seeds}), as on test\_website, while the 8B writes about a fifth of the steps and the pipeline runs faster than the 8B alone (\cref{tab:s_m2w_task}; speed: \cref{tab:speed}).

\begin{table}[!htb]
\centering
\small
\setlength{\tabcolsep}{2.5pt}
\begin{tabular}{@{}lcc@{}}
\toprule
Method & M2W task & M2W web \\
\midrule
Target model (8B) & 65.7 / 1.00 & 58.0 / 1.00 \\
8B + EAGLE-3 & 67.0 / 1.55 & 59.7 / 1.61 \\
\rowcolor{vgmist}\method{} & \textbf{71.3} / 1.17 & \textbf{64.0} / 1.15 \\
\quad + EAGLE-3 on both & 70.0 / 1.50 & 62.0 / 1.52 \\
\quad\quad projected, $D$ at 0.16\,s & \emph{70.0 / 1.86} & \emph{62.0 / 1.84} \\
\bottomrule
\end{tabular}
\caption{\textbf{\method{} is the most accurate on both Mind2Web subsets, and with EAGLE-3 it is nearly as fast as EAGLE-3 alone.} Step success (\%) / mean speedup over the 8B, measured end to end at batch size 1 on the same test steps (300 each). EAGLE-3 uses the public Qwen3-VL heads with five speculative tokens and is exact only up to floating-point rounding, so the 8B's step success with it differs slightly from that of the plain 8B. Bold: most accurate. The last row (italic) is projected, with the measured API latency of $D$ (0.4--0.6\,s per step) replaced by its nominal 0.16\,s. Thresholds: 0.41 on test\_task (an operating point of the development sweep), 0.5 on test\_website. AndroidControl timings: \cref{tab:s_components,tab:s_cost}.}
\label{tab:speed}
\end{table}
\FloatBarrier

\clearpage
\subsection{Full Agent Results}
\label{sec:s_agents}
\begin{figure}[!htb]
\centering
\includegraphics[width=\linewidth]{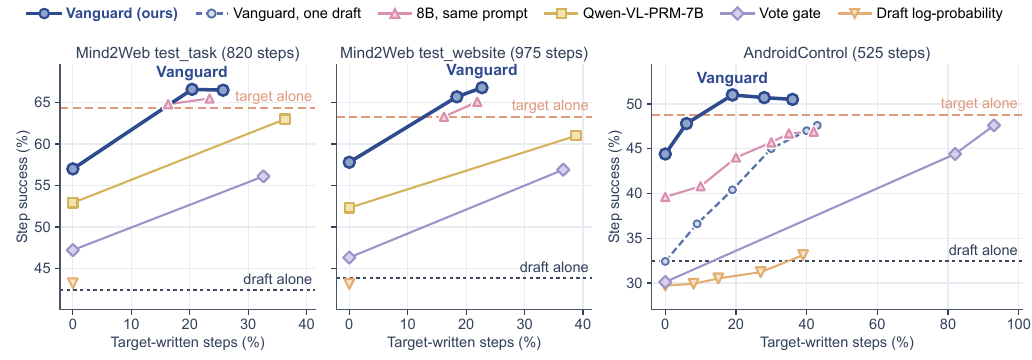}
\caption{\textbf{On all three agent test sets, \method{} reaches the highest step success of any configuration shown, above the 8B alone, while the 8B writes at most a quarter of the steps.} Higher and further left is better. Step success against the share of target-written steps for the four-candidate pipelines with fallback, one line per judge through every threshold of its sweep (points at 0\%: no fallback), on Mind2Web test\_task (820 steps), Mind2Web test\_website (975 steps) and AndroidControl (525 steps). On AndroidControl, the dashed light-navy line is \method{} with one draft per step. Dashed horizontal lines: the 8B target alone and the 2B draft alone. Full numbers: \cref{tab:s_m2w_task,tab:s_m2w_web,tab:s_android}.}
\label{fig:s_agent_sweeps}
\end{figure}

\Cref{tab:s_m2w_task,tab:s_m2w_web,tab:s_android} list the configurations on the agent test sets, including single-draft pipelines and full threshold sweeps.
\Cref{fig:s_agent_sweeps} draws the sweeps.

\paragraph{Standard Mind2Web Metrics.}
In the standard metrics of Mind2Web (\cref{tab:s_m2w_metrics}), \method{} selects the right element more often than the 8B, and more tasks have every evaluated step correct.

\begin{table}[!htb]
\centering
\small
\setlength{\tabcolsep}{5pt}
\begin{tabular}{@{}lcc@{}}
\toprule
Method & Element & Task \\
\midrule
Target model alone (8B) & 74.1 & 18.0 \\
\rowcolor{vgmist}\method{} & \textbf{78.4} & \textbf{19.9} \\
\bottomrule
\end{tabular}
\caption{\textbf{\method{} selects the right element more often and completes more tasks than the 8B.} Standard Mind2Web metrics (\%) over both test sets: element accuracy and task success, which counts tasks whose every evaluated step is correct.}
\label{tab:s_m2w_metrics}
\end{table}

\FloatBarrier

In every panel, the line of \method{} reaches the highest step success of all judges while the 8B writes at most a quarter of the steps (\cref{fig:s_agent_sweeps}).
\FloatBarrier

\begin{table}[!htb]
\centering
\footnotesize
\setlength{\tabcolsep}{5pt}
\begin{tabular}{@{}lcccc@{}}
\toprule
Configuration & $\tau$ & Step success (\%) $\uparrow$ & Target steps (\%) $\downarrow$ & Wins/losses vs.\ 8B ($p$) \\
\midrule
Draft model alone (2B) & -- & 42.4 & 0 & 48/228, $<$0.001 \\
Target model alone (8B) & -- & 64.4 & 100 & -- \\
Four candidates, random & none & 45.0 & 0.0 & 60/219, $<$0.001 \\
Four candidates, vote gate & none & 47.2 & 0.0 & 58/199, $<$0.001 \\
Four candidates, vote gate & 0.75 & 56.1 & 32.6 & 37/105, $<$0.001 \\
Four candidates, log-probability & none & 43.2 & 0.0 & 46/220, $<$0.001 \\
Four candidates, 8B same prompt & 0.5 & 64.8 & 16.3 & 69/66, 0.863 \\
Four candidates, 8B same prompt & 0.9399 & 65.5 & 23.4 & 63/54, 0.460 \\
Four candidates, Qwen-VL-PRM-7B & none & 52.9 & 0.0 & 74/168, $<$0.001 \\
Four candidates, Qwen-VL-PRM-7B & 0.7773 & 63.0 & 36.3 & 51/62, 0.347 \\
\rowcolor{vgmist}Four candidates, \method{} & none & 57.0 & 0.0 & 74/135, $<$0.001 \\
\rowcolor{vgmist}Four candidates, \method{} & 0.41 & \textbf{66.6} & 20.4 & 69/51, 0.120 \\
\rowcolor{vgmist}Four candidates, \method{} & 0.5 & 66.5 & 25.7 & 61/44, 0.118 \\
One draft, 8B same prompt & 0.5 & 60.1 & 35.0 & 42/77, 0.002 \\
\rowcolor{vgmist}One draft, \method{} & 0.5 & 63.3 & 44.1 & 36/45, 0.374 \\
\bottomrule
\end{tabular}
\caption{\textbf{On test\_task, \method{} is the most accurate configuration, above the 8B alone at both of its thresholds.} Mind2Web with reasoning, test\_task test part (820 steps); thresholds fixed on the development part.}
\label{tab:s_m2w_task}
\end{table}

\FloatBarrier
On test\_task, both thresholds of \method{} with four candidates and fallback lie above the 8B alone, and no configuration in \cref{tab:s_m2w_task} is more accurate.
\FloatBarrier

\begin{table}[!htb]
\centering
\footnotesize
\setlength{\tabcolsep}{5pt}
\begin{tabular}{@{}lcccc@{}}
\toprule
Configuration & $\tau$ & Step success (\%) $\uparrow$ & Target steps (\%) $\downarrow$ & Wins/losses vs.\ 8B ($p$) \\
\midrule
Draft model alone (2B) & -- & 43.8 & 0 & 73/263, $<$0.001 \\
Target model alone (8B) & -- & 63.3 & 100 & -- \\
Four candidates, random & none & 42.6 & 0.0 & 88/290, $<$0.001 \\
Four candidates, vote gate & none & 46.3 & 0.0 & 85/251, $<$0.001 \\
Four candidates, vote gate & 0.75 & 56.9 & 36.6 & 56/118, $<$0.001 \\
Four candidates, log-probability & none & 43.1 & 0.0 & 83/280, $<$0.001 \\
Four candidates, 8B same prompt & 0.5 & 63.3 & 16.2 & 95/95, 1.000 \\
Four candidates, 8B same prompt & 0.9399 & 65.1 & 21.9 & 90/72, 0.181 \\
Four candidates, Qwen-VL-PRM-7B & none & 52.3 & 0.0 & 105/212, $<$0.001 \\
Four candidates, Qwen-VL-PRM-7B & 0.7773 & 61.0 & 38.8 & 68/90, 0.094 \\
\rowcolor{vgmist}Four candidates, \method{} & none & 57.8 & 0.0 & 106/159, 0.001 \\
\rowcolor{vgmist}Four candidates, \method{} & 0.41 & 65.7 & 18.4 & 98/74, 0.079 \\
\rowcolor{vgmist}Four candidates, \method{} & 0.5 & \textbf{66.8} & 22.7 & 95/61, 0.008 \\
One draft, 8B same prompt & 0.5 & 60.8 & 33.2 & 67/91, 0.067 \\
\rowcolor{vgmist}One draft, \method{} & 0.5 & 64.4 & 41.3 & 67/56, 0.367 \\
\bottomrule
\end{tabular}
\caption{\textbf{On the unseen websites, \method{} is the most accurate configuration, significantly above the 8B alone at $\tau = 0.5$.} Mind2Web with reasoning, test\_website (975 steps, unseen websites), with the thresholds of test\_task.}
\label{tab:s_m2w_web}
\end{table}

\FloatBarrier
The same holds on the unseen websites of \cref{tab:s_m2w_web}; the lower threshold, $\tau = 0.41$, hands fewer steps to the 8B (18.4\% of the steps, against 22.7\% at $\tau = 0.5$).
\FloatBarrier

\begin{table}[!htb]
\centering
\scriptsize
\setlength{\tabcolsep}{2.5pt}
\renewcommand{\arraystretch}{0.92}
\begin{tabular}{@{}lccccc@{}}
\toprule
Configuration & $\tau$ & Step success (\%) $\uparrow$ & Target steps (\%) $\downarrow$ & Speedup over 8B $\uparrow$ & Wins/losses vs.\ 8B ($p$) \\
\midrule
Draft model alone (2B) & -- & 32.4 & 0 & 2.16$\times$ & 22/108, $<$0.01 \\
Target model alone (8B) & -- & 48.8 & 100 & 1.00$\times$ & -- \\
Four candidates, random & none & 32.4 & 0 & 1.59$\times$ & 38/124, $<$0.01 \\
Four candidates, vote gate & none & 30.1 & 0 & 1.59$\times$ & 29/127, $<$0.01 \\
Four candidates, vote gate & 0.75 & 44.4 & 82 & 0.69$\times$ & 5/28, $<$0.01 \\
Four candidates, vote gate & 1.0 & 47.6 & 93 & 0.64$\times$ & 3/9, 0.15 \\
Four candidates, log-probability & none & 29.7 & 0 & 1.59$\times$ & 31/131, $<$0.01 \\
Four candidates, log-probability & -0.2907 & 29.9 & 8 & 1.40$\times$ & 27/126, $<$0.01 \\
Four candidates, log-probability & -0.2516 & 30.5 & 15 & 1.29$\times$ & 24/120, $<$0.01 \\
Four candidates, log-probability & -0.2181 & 31.2 & 27 & 1.11$\times$ & 20/112, $<$0.01 \\
Four candidates, log-probability & -0.1947 & 33.1 & 39 & 0.98$\times$ & 17/99, $<$0.01 \\
Four candidates, 8B same prompt & none & 39.6 & 0 & 1.09$\times$ & 38/86, $<$0.01 \\
Four candidates, 8B same prompt & 0.0052 & 40.8 & 10 & 0.99$\times$ & 32/74, $<$0.01 \\
Four candidates, 8B same prompt & 0.04743 & 44.0 & 20 & 0.89$\times$ & 28/53, 0.01 \\
Four candidates, 8B same prompt & 0.2504 & 45.7 & 30 & 0.82$\times$ & 25/41, 0.06 \\
Four candidates, 8B same prompt & 0.5 & 46.7 & 35 & 0.78$\times$ & 23/34, 0.18 \\
Four candidates, 8B same prompt & 0.8095 & 46.9 & 42 & 0.74$\times$ & 20/30, 0.20 \\
\rowcolor{vgmist}Four candidates, \method{} & none & 44.4 & 0 & 1.26$\times$ & 48/71, 0.04 \\
\rowcolor{vgmist}Four candidates, \method{} & 0.314 & 47.8 & 6 & 1.16$\times$ & 47/52, 0.69 \\
\rowcolor{vgmist}Four candidates, \method{} & 0.494 & \textbf{51.0} & 19 & 1.01$\times$ & 43/31, 0.20 \\
\rowcolor{vgmist}Four candidates, \method{} & 0.5 & \textbf{51.0} & 19 & 1.01$\times$ & 43/31, 0.20 \\
\rowcolor{vgmist}Four candidates, \method{} & 0.62 & 50.7 & 28 & 0.92$\times$ & 37/27, 0.26 \\
\rowcolor{vgmist}Four candidates, \method{} & 0.69 & 50.5 & 36 & 0.86$\times$ & 31/22, 0.27 \\
\rowcolor{vgmist}Four candidates, \method{}$^\dagger$ & none & 44.4 & 0 & 1.45$\times$ & 48/71, 0.04 \\
\rowcolor{vgmist}Four candidates, \method{}$^\dagger$ & 0.314 & 47.8 & 6 & 1.32$\times$ & 47/52, 0.69 \\
\rowcolor{vgmist}Four candidates, \method{}$^\dagger$ & 0.494 & \textbf{51.0} & 19 & 1.13$\times$ & 43/31, 0.20 \\
\rowcolor{vgmist}Four candidates, \method{}$^\dagger$ & 0.5 & \textbf{51.0} & 19 & 1.13$\times$ & 43/31, 0.20 \\
\rowcolor{vgmist}Four candidates, \method{}$^\dagger$ & 0.62 & 50.7 & 28 & 1.02$\times$ & 37/27, 0.26 \\
\rowcolor{vgmist}Four candidates, \method{}$^\dagger$ & 0.69 & 50.5 & 36 & 0.95$\times$ & 31/22, 0.27 \\
One draft, log-probability & none & 32.4 & 0 & 2.16$\times$ & 22/108, $<$0.01 \\
One draft, log-probability & -0.2765 & 32.4 & 11 & 1.72$\times$ & 15/101, $<$0.01 \\
One draft, log-probability & -0.2425 & 33.7 & 18 & 1.54$\times$ & 15/94, $<$0.01 \\
One draft, log-probability & -0.2174 & 34.5 & 26 & 1.37$\times$ & 13/88, $<$0.01 \\
One draft, log-probability & -0.2017 & 35.6 & 35 & 1.23$\times$ & 12/81, $<$0.01 \\
One draft, 8B same prompt & none & 32.4 & 0 & 1.33$\times$ & 22/108, $<$0.01 \\
One draft, 8B same prompt & 0.0001 & 34.1 & 7 & 1.22$\times$ & 21/98, $<$0.01 \\
One draft, 8B same prompt & 0.0032 & 39.2 & 21 & 1.04$\times$ & 20/70, $<$0.01 \\
One draft, 8B same prompt & 0.018 & 41.7 & 32 & 0.93$\times$ & 18/55, $<$0.01 \\
One draft, 8B same prompt & 0.0915 & 45.0 & 42 & 0.85$\times$ & 15/35, 0.01 \\
One draft, 8B same prompt & 0.5 & 46.5 & 53 & 0.78$\times$ & 13/25, 0.07 \\
\rowcolor{vgmist}One draft, \method{} & none & 32.4 & 0 & 1.58$\times$ & 22/108, $<$0.01 \\
\rowcolor{vgmist}One draft, \method{} & 0.1 & 36.6 & 9 & 1.39$\times$ & 22/86, $<$0.01 \\
\rowcolor{vgmist}One draft, \method{} & 0.204 & 40.4 & 19 & 1.22$\times$ & 19/63, $<$0.01 \\
\rowcolor{vgmist}One draft, \method{} & 0.346 & 45.0 & 30 & 1.07$\times$ & 19/39, 0.01 \\
\rowcolor{vgmist}One draft, \method{} & 0.47 & 47.0 & 40 & 0.96$\times$ & 16/25, 0.21 \\
\rowcolor{vgmist}One draft, \method{} & 0.5 & 47.6 & 43 & 0.94$\times$ & 16/22, 0.42 \\
\rowcolor{vgmist}One draft, \method{}$^\dagger$ & none & 32.4 & 0 & 1.89$\times$ & 22/108, $<$0.01 \\
\rowcolor{vgmist}One draft, \method{}$^\dagger$ & 0.1 & 36.6 & 9 & 1.62$\times$ & 22/86, $<$0.01 \\
\rowcolor{vgmist}One draft, \method{}$^\dagger$ & 0.204 & 40.4 & 19 & 1.40$\times$ & 19/63, $<$0.01 \\
\rowcolor{vgmist}One draft, \method{}$^\dagger$ & 0.346 & 45.0 & 30 & 1.20$\times$ & 19/39, 0.01 \\
\rowcolor{vgmist}One draft, \method{}$^\dagger$ & 0.47 & 47.0 & 40 & 1.07$\times$ & 16/25, 0.21 \\
\rowcolor{vgmist}One draft, \method{}$^\dagger$ & 0.5 & 47.6 & 43 & 1.04$\times$ & 16/22, 0.42 \\
\bottomrule
\end{tabular}
\caption{\textbf{On AndroidControl, \method{} is the most accurate configuration and stays above the 8B alone at every threshold that sends 19--36\% of the steps to it.} AndroidControl with reasoning, test part (525 steps), full threshold sweeps. Speedups in this table are estimated over all 525 steps from the per-component times of \cref{tab:s_components} ($c_K + c_D + (1-\alpha_\tau)\,c_p$ against $c_p$); the end-to-end measurement on the timed subset of these steps is 1.06$\times$ at $\tau = 0.5$ (\cref{tab:s_cost}). $^\dagger$: speedup with the decision model at its nominal latency of 0.16\,s instead of the measured latency.}
\label{tab:s_android}
\end{table}

On AndroidControl (\cref{tab:s_android}), \method{} stays above the 8B alone at every threshold that sends 19--36\% of the steps to it. With one draft per step, \method{} and the 8B asked the same question need far more target-written steps to reach a similar accuracy, which is why \method{} draws four candidates.

\clearpage
\subsection{An Embodied Benchmark: EB-ALFRED}
\label{sec:s_alfred}

To test \method{} on a third kind of agent, we use EB-ALFRED, the household-robot environment of EmbodiedBench~\cite{yang2025embodiedbench} built on ALFRED~\cite{shridhar2020alfred}: the agent receives a language instruction and acts from egocentric images with high-level actions (navigate to, pick up, put down, open, close, turn on, turn off, slice) over the objects of the scene.
We evaluate offline on the trajectory dataset released with the benchmark: for each of the 256 episodes (of 300) that have a successful trajectory, we take the shortest one as the gold path and keep its successfully executed actions, which gives 2,607 steps, split by episode into 758 development and 1,849 test steps.
At each step the agent sees the current image, the instruction and the gold history, writes two to four sentences of reasoning and chooses the next action among eight options: the gold action and seven distractors, drawn first from actions whose preconditions provably fail in the current state given the executed history (picking up while holding something, putting down with empty hands, opening what is already open, \dots), then from actions with the same verb as the gold action, then at random from the episode's action list, so that a model must track the state of the task to answer.
The judge is shown the image, the instruction, the executed actions and the chosen action, but not the reasoning; thresholds are chosen on the development steps as in the main text.

\Cref{tab:s_alfred} shows the same picture as on the web and mobile tasks.
With both families, \method{} is significantly more accurate than the target model alone while the target writes about a quarter of the steps, and no other judge of the same candidates comes close at a similar share; with Qwen3-VL, the draft's self-check and the target asked the same question only approach the target model when they hand it two thirds or more of the steps.
With Qwen2.5-VL, the 3B draft alone already scores 69.7\% against the 7B target's 67.9\%, and its four samples contain a correct action on 83\% of the steps; \method{} turns this into a gain of 8.4 points over the target.
Timing on this task resembles AndroidControl.
Measured end to end at batch size 1 on the first 300 test steps (each model on a GPU of its own), four 2B candidates take 0.97\,s, the decision model 0.47\,s and an 8B step 2.09\,s, so the condition of the main text holds: \method{} is 1.06$\times$ faster than the 8B on average and 1.39$\times$ at the median, while scoring 84.7\% against 78.7\% on these steps (25/7, $p = 0.002$).
With EAGLE-3 on the fallback it is 1.19$\times$ faster on average (1.44$\times$ at the nominal latency of $D$), against 1.63$\times$ for EAGLE-3 alone, and more accurate than EAGLE-3 alone (83.3\% against 77.7\%, 23/6, $p = 0.002$).

\begin{table}[!htb]
\centering
\footnotesize
\setlength{\tabcolsep}{1.8pt}
\begin{tabular}{@{}lcccc@{}}
\toprule
& \multicolumn{2}{c}{Qwen3-VL (2B, 8B)} & \multicolumn{2}{c}{Qwen2.5-VL (3B, 7B)} \\
\cmidrule(lr){2-3}\cmidrule(l){4-5}
Method & Success & Target & Success & Target \\
\midrule
Draft model alone & 60.6 & 0 & 69.7 & 0 \\
Target model alone & 74.2 & 100 & 67.9 & 100 \\
Oracle (pass@4) & 76.3 & 0 & 83.2 & 0 \\
\midrule
Random of four & 61.8 & 0 & 62.5 & 0 \\
Vote gate & 72.7 & 39 & 71.3 & 27 \\
Draft model, self-check & 72.8 & 75 & 72.9 & 24 \\
Target model, same prompt & 74.7 & 66 & 70.3 & 78 \\
\rowcolor{vgmist}\method{} & \textbf{77.4} & 28 & \textbf{76.3} & 27 \\
\bottomrule
\end{tabular}
\caption{\textbf{With both families, \method{} beats the target on EB-ALFRED while the target writes about a quarter of the steps.} Step success (\%) on the 1,849 test steps with four draft candidates per step and fallback to the target model at thresholds fixed on development data; ``Target'' is the share of target-written steps (\%), and each judge is shown at its development-chosen threshold. \method{} against the target model alone: 132/73 steps ($p = 5\times10^{-5}$) with Qwen3-VL and 213/58 ($p = 5\times10^{-22}$) with Qwen2.5-VL. With Qwen3-VL, the gate AUROC of the selected candidate's score is 0.859 for \method{}, 0.631 for the draft's self-check and 0.666 for the target asked the same question; with thresholds matched on development data to the share of \method{}, those two judges reach 66.1 (22\% target-written) and 70.8 (30\%).}
\label{tab:s_alfred}
\end{table}

\clearpage
\subsection{Closed-Loop Episodes}
\label{sec:s_closed_loop}

\begin{table}[!htb]
\centering
\footnotesize
\setlength{\tabcolsep}{3pt}
\begin{tabular}{@{}lcccc@{}}
\toprule
& \multicolumn{2}{c}{GoToLocal} & \multicolumn{2}{c}{PickupLoc} \\
\cmidrule(lr){2-3}\cmidrule(l){4-5}
Method & Success & Steps & Success & Steps \\
\midrule
Draft model alone (2B) & 37 & 33.6 & 16 & 40.1 \\
Target model alone (8B) & 61 & 22.7 & 36 & 32.0 \\
Target model, same prompt & 63 & 19.4 & 53 & 25.6 \\
\rowcolor{vgmist}\method{} & \textbf{86} & 13.1 & \textbf{65} & 20.7 \\
\bottomrule
\end{tabular}
\caption{\textbf{In closed loop, \method{} completes far more episodes than the 8B while the 8B writes a quarter of the decisions.} Episode success (\%) and mean number of environment steps on 100 test episodes (seeds 30--129) of two BabyAI tasks; thresholds chosen on 30 development episodes ($\tau = 0.3$ for both judges on both tasks). \method{} against the 8B alone: 27/2 episodes on GoToLocal ($p = 2\times10^{-6}$) and 34/5 on PickupLoc ($p = 2\times10^{-6}$), with the 8B writing 25\% of the decisions on both. The 8B asked the same question as a judge hands 83--90\% of the decisions back to the 8B.}
\label{tab:s_loop}
\end{table}

The agent benchmarks above score each step given the gold history.
To see whether the gains carry over to full episodes, where the agent's own errors accumulate, we run \method{} in the loop on two BabyAI tasks~\cite{chevalierboisvert2019babyai} of Minigrid~\cite{chevalierboisvert2023minigrid}: GoToLocal (reach the object named in the mission in a room with distractors) and PickupLoc (pick up the object described by its position).
The agent sees the whole room rendered with column letters and row numbers, in the style of Set-of-Mark, and acts with absolute moves (north, south, east, west), pick up, drop and toggle; on PickupLoc, the prompt of every method also states what is in the cell in front of the agent and what it carries; an episode succeeds when the mission is completed within the environment's step limit.
At every decision the 2B draft samples four actions with a short reasoning, the judge scores each distinct action against the rendered room, and the 8B writes the action when the highest score is below $\tau$; the 2B and the 8B alone act greedily from the same prompt.
On GoToLocal (\cref{tab:s_loop}), \method{} completes 86\% of the test episodes against 61\% for the 8B alone and 37\% for the 2B alone, while the 8B writes a quarter of the decisions; the 8B asked the same question as a judge accepts almost no draft and ends where the 8B alone does.
On PickupLoc, where the agent must also pick the object up, the 8B alone completes 36\% of the test episodes and \method{} 65\% (34/5 episodes, $p = 2\times10^{-6}$), again with a quarter of the decisions written by the 8B; the 8B as judge reaches 53\% while writing 83\% of the decisions itself, and \method{} stays ahead of it, as on GoToLocal (25/2 episodes there; \cref{tab:s_pairs}).
What the closed-loop runs show is that choosing among drafts and falling back on the steps the judge rejects raises episode success rather than merely step success.

\begin{table}[!htb]
\centering
\footnotesize
\setlength{\tabcolsep}{3pt}
\begin{tabular}{@{}lcccccc@{}}
\toprule
Method & Success & Steps & 8B steps & s / step & s / episode & TFLOP / step \\
\midrule
Draft model alone (2B) & 12.1 & 16.5 & 0\% & 1.57 / 1.20 & 25.8 & 5.6 \\
Target model alone (8B) & 28.0 & 13.9 & 100\% & 2.99 / 2.76 & 41.4 & 21.6 \\
\rowcolor{vgmist}\method{} & \textbf{29.3} & 14.9 & 29\% & 2.79 / 1.83 & 41.7 & 13.7 \\
\bottomrule
\end{tabular}
\caption{\textbf{On AndroidWorld, \method{} matches the 8B while the 8B writes 29\% of the steps.} Task success (\%) on all 116 tasks (one parameter draw each, identical across methods), mean steps per episode, and the share of steps written by the 8B. Time per step (mean / median) and per episode is model and judge time estimated from the logged tokens (see text); compute counts $2\times$ parameters $\times$ tokens of the two models. \method{} against the 8B alone: 12/11 tasks; the 2B alone against the 8B: 6/25 ($p = 9\times10^{-4}$).}
\label{tab:s_androidworld}
\end{table}

Beyond the BabyAI episodes of the main text, we also ran the three agents in the loop on AndroidWorld~\cite{rawles2024androidworld}, whose 116 tasks run in real apps on an Android emulator and are scored by the benchmark's own checkers. The prompt, action format, judge question and $\tau = 0.5$ are those of AndroidControl, not tuned on AndroidWorld; each episode may take up to ten times the task's complexity rating in the benchmark, in steps.
\method{} succeeds on 29.3\% of the tasks against 28.0\% for the 8B alone and 12.1\% for the 2B alone (\cref{tab:s_androidworld}), with the 8B writing 29\% of the steps: it matches the 8B (higher on 12 tasks, lower on 11), whereas the 2B alone is far below it.
The emulator runs on a separate machine that reaches the models over the network, so measured wall time is not comparable with the timing in the main text; we therefore estimate model time from the logged tokens with the batch-1 rates of our timing runs on the same GPUs (the 8B: 0.28~s plus 24~ms per output token; the four draft samples: 13~ms per output token per sample) and add the judge latency measured on that machine (0.30~s per step).
Per step, \method{} is 1.07$\times$ faster than the 8B on average and 1.51$\times$ at the median, and uses 0.63$\times$ its compute; per episode it is as fast as the 8B (0.99$\times$), because its episodes are one step longer on average. The decision model cost \$0.47 for the 116 episodes.

\FloatBarrier
\clearpage
\subsection{Two More Model Families}
\label{sec:s_families}

\begin{table}[!htb]
\centering
\footnotesize
\setlength{\tabcolsep}{2.5pt}
\begin{tabular}{@{}lcccc@{}}
\toprule
& \multicolumn{2}{c}{Mind2Web} & \multicolumn{2}{c}{EB-ALFRED} \\
\cmidrule(lr){2-3}\cmidrule(l){4-5}
Method & Success & Target & Success & Target \\
\midrule
\multicolumn{5}{@{}l}{\emph{Qwen2.5-VL~\cite{bai2025qwen25vl}, 3B draft, 7B target}} \\
Draft alone & 39.6 & 0 & 69.7 & 0 \\
Target alone & 50.5 & 100 & 67.9 & 100 \\
Oracle (pass@4) & 59.4 & 0 & 83.2 & 0 \\
Target, same prompt & 53.2 & 43 & 70.3 & 78 \\
\rowcolor{vgmist}\method{} & \textbf{58.6} & 20 & \textbf{76.3} & 27 \\
\midrule
\multicolumn{5}{@{}l}{\emph{Qwen3.5~\cite{qwen2026qwen35}, 2B draft, 9B target}} \\
Draft alone & 44.2 & 0 & 50.1 & 0 \\
Target alone & 66.0 & 100 & 72.3 & 100 \\
Oracle (pass@4) & 67.5 & 0 & 73.4 & 0 \\
Target, same prompt & 63.7 & 30 & 72.5 & 58 \\
\rowcolor{vgmist}\method{} & \textbf{66.5} & 19 & \textbf{73.2} & 40 \\
\bottomrule
\end{tabular}
\caption{\textbf{The gain carries over to two more model families, most of all to Qwen2.5-VL, whose draft has the larger headroom.} Step success (\%) and share of target-written steps (\%) with two other model families, same prompts, splits and protocol as the main text (Mind2Web: both test sets pooled, 1,795 steps; EB-ALFRED: 1,849); thresholds chosen on development data, the target asked the same question at its development-chosen threshold. \method{} against the target alone, Qwen2.5-VL: 275/130 ($p = 5\times10^{-13}$) and 213/58 ($p = 5\times10^{-22}$).}
\label{tab:s_families}
\end{table}

\Cref{tab:s_families} repeats the main experiment with two further families, with nothing changed but the models.
With Qwen2.5-VL, whose 3B draft produces a correct action among its four samples more often than the 7B target answers correctly on Mind2Web and EB-ALFRED, \method{} is significantly more accurate than the target on both, by 8.1 and 8.4 points.
With Qwen3.5, whose 2B draft reaches a pass@4 close to the 9B's own accuracy on Mind2Web and EB-ALFRED (67.5 against 66.0, 73.4 against 72.3), \method{} is above the target on both (+0.5 and +0.9 points) while the 9B writes 19 and 40\% of the steps; the judge still matters there, since the 9B asked the same question falls significantly below the 9B alone on Mind2Web at its own threshold (102/142, $p = 0.01$). At its own development threshold, the target asked the same question is below \method{} in all four cells of \cref{tab:s_families}, and in every one it writes more of the steps itself (78 against 27\% on EB-ALFRED with Qwen2.5-VL, for example).

\clearpage
\subsection{Cross-Family Targets}
\label{sec:s_cross}

\begin{table}[!htb]
\centering
\footnotesize
\setlength{\tabcolsep}{3pt}
\begin{tabular}{@{}lcccc@{}}
\toprule
Target & Alone & Oracle (2B pass@4) & \method{} & W/L, $p$ \\
\midrule
\multicolumn{5}{@{}l}{\emph{Gemma-3-27B}} \\
Mind2Web & 64.1 & 62.8 & \textbf{66.6} (24) & 169/123, 0.008 \\
AndroidControl & 34.1 & 52.8 & \textbf{45.9} (6) & 125/63, $7\times10^{-6}$ \\
EB-ALFRED & 76.6 & 76.3 & \textbf{78.3} (32) & 112/80, 0.025 \\
\midrule
\multicolumn{5}{@{}l}{\emph{Gemma-3-12B}} \\
Mind2Web test\_task & 59.6 & 62.3 & \textbf{64.0} (26) & 78/42, 0.001 \\
\bottomrule
\end{tabular}
\caption{\textbf{With a Gemma-3 target, \method{} significantly surpasses the target on every benchmark.} The Qwen3-VL-2B draft and \method{} with a target from another family (step success, \%; ``Alone'': the target model alone; share of target-written steps, \%, in parentheses; W/L and $p$ against the target alone). The draft's four samples and their scores are those of the main experiments; only the target changes, with the same prompts and splits and a threshold chosen on each pair's development data (Mind2Web: both test sets, 1,795 steps, except Gemma-3-12B, which we ran on test\_task only).}
\label{tab:s_cross}
\end{table}

Step-level speculation exchanges whole actions, not tokens, so the draft and the target need not share a tokenizer or a family.
\Cref{tab:s_cross} pairs the Qwen3-VL-2B draft with Gemma-3 targets.
With Gemma-3-27B, \method{} is significantly more accurate than the target on all three benchmarks, while the 27B writes 6--32\% of the steps; with Gemma-3-12B it is 4.4 points above the target on Mind2Web.
The judge matters here as well: with the same candidates and the Gemma-3-27B fallback, the process reward model Qwen-VL-PRM-7B at the same share of fallbacks scores 60.2\%, 22.9\% and 70.5\%, significantly below both \method{} (195/79, 135/14 and 225/80 steps; $p \le 2\times10^{-12}$) and the 27B alone ($p \le 3\times10^{-4}$).
In the terms of \cref{sec:s_passk}, the 2B's headroom against the 27B is close to zero on Mind2Web and EB-ALFRED (1.3 and 0.3 points below) and large on AndroidControl, where the gain is also largest.
Gemma-3-27B itself asked the same question is a stronger judge than the PRM but still falls short: at the share of fallbacks of \method{} it scores 63.0\%, 37.7\% and 74.4\%, significantly below \method{} on all three benchmarks (118/53, 69/26 and 131/58 steps; $p \le 2\times10^{-5}$).
The cross-family pairing is also where step-level speculation saves the most time, because the 27B makes each fallback expensive.
Measured end to end at batch size 1 (2B on one GPU, the 27B on two, no caching; $\tau = 0.41$, 0.5 and 0.451, the thresholds of the 8B timing runs), \method{} is 1.34$\times$, 1.33$\times$ and 1.36$\times$ faster than the 27B alone on the first 300, 200 and 300 test steps of Mind2Web, AndroidControl and EB-ALFRED (1.91--1.99$\times$ at the median; 1.45--1.60$\times$ at the nominal latency of the decision model), while being more accurate on the same steps (67.7\% against 65.0\%, 45.0\% against 36.0\%, and 83.7\% against 78.3\%). On the full Mind2Web test sets, the same pair is significantly more accurate than the 27B (169/123 steps; \cref{tab:s_cross}), and at the timed thresholds it scores 46.3\% on the full AndroidControl test set ($\tau = 0.5$, 19\% target-written steps; 116/52 steps against the 27B, $p = 9\times10^{-7}$) and 78.2\% against 76.6\% on the full EB-ALFRED test set ($\tau = 0.451$, 28\% target-written steps; 124/94, $p = 0.049$).

\begin{figure}[!htb]
\centering
\includegraphics[width=\linewidth]{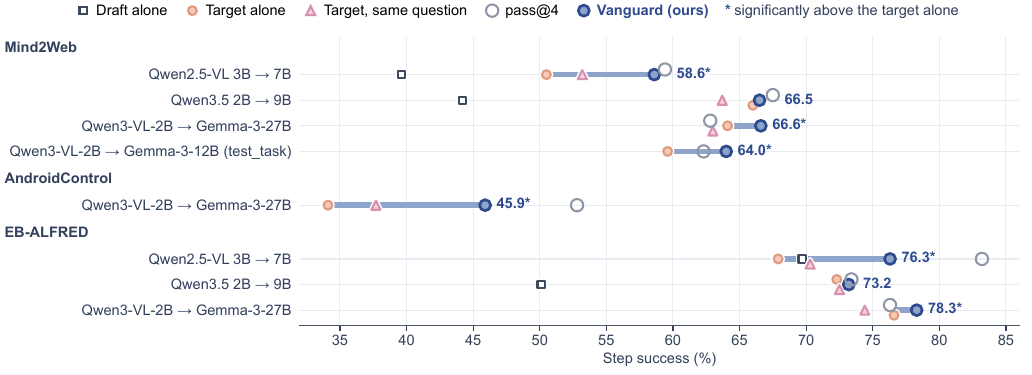}
\caption{\textbf{With four further draft--target pairs, including Gemma-3 targets from another family, \method{} is above the target alone in every pair and on every benchmark shown.} Higher is better. Step success (\%) per (draft$\to$target pair, benchmark): the draft alone (white square), the target alone (peach), the target asked the same question (pink), the draft's pass@4 (grey ring) and \method{} (navy, value printed); the bar joins the target alone and \method{}. Mind2Web pools both test sets except for Gemma-3-12B (test\_task). $^*$: \method{} significantly above the target alone. Full numbers: \cref{tab:s_families,tab:s_cross}; Gemma-3-27B asked the same question: \cref{sec:s_cross}.}
\label{fig:s_families}
\end{figure}

\Cref{fig:s_families} sets the pairs of \cref{tab:s_families,tab:s_cross} side by side.

\paragraph{A Non-Qwen Pair.}
We also ran Gemma-3-12B as a draft for Gemma-3-27B; on the development steps of Mind2Web, AndroidControl and EB-ALFRED, its pass@4 is above the 27B by 7.4, 11.3 and 1.8 points.
On the test steps, where its pass@4 is 71.1, 40.6 and 78.0\%, \method{} with the 12B draft is significantly more accurate than the 27B on Mind2Web (69.1\% against 64.1\%, 190/99, $p = 9\times10^{-8}$; 12\% target-written steps) and AndroidControl (39.4\% against 34.1\%, 47/19, $p = 8\times10^{-4}$), and it is also ahead on EB-ALFRED (77.7\% against 76.6\%).

\clearpage
\subsection{Judges on Every Family and Benchmark}
\label{sec:s_judges_all}

\begin{table}[!htb]
\centering
\footnotesize
\setlength{\tabcolsep}{2.5pt}
\begin{tabular}{@{}llccc@{}}
\toprule
Family & Benchmark & \method{} & PRM & Target, same q. \\
\midrule
Qwen3-VL & Mind2Web & \textbf{66.2} (31) & 61.2 & 65.8 \\
 & AndroidControl & \textbf{50.7} (28) & 30.9 & 45.9 \\
 & EB-ALFRED & \textbf{77.4} (28) & 69.6 & 70.8 \\
Qwen3.5 & Mind2Web & \textbf{66.5} (19) & 57.8 & 62.8 \\
 & AndroidControl & \textbf{43.8} (30) & 26.9 & 41.3 \\
 & EB-ALFRED & \textbf{73.2} (40) & 63.7 & 70.5 \\
Qwen2.5-VL & Mind2Web & \textbf{58.6} (20) & 50.1 & 51.6 \\
 & AndroidControl & \textbf{36.2} (34) & 26.5 & 25.9 \\
 & EB-ALFRED & \textbf{76.3} (27) & 66.8 & 67.6 \\
\bottomrule
\end{tabular}
\caption{\textbf{\method{} is the most accurate judge in every (family, benchmark) cell.} Three judges of the same draft candidates in each (family, benchmark) cell we report (step success, \%, on the test steps). Bold: the most accurate judge in the cell. \method{} uses its threshold chosen on development data (share of target-written steps, \%, in parentheses); the vision-language process reward model Qwen-VL-PRM-7B and the target model asked the same question use thresholds rate-matched on development data to that share. \method{} is significantly more accurate than the PRM in all nine cells ($p \le 4\times10^{-7}$) and than the target asked the same question in seven ($p \le 0.002$). For Qwen3-VL, this table chooses every threshold from its own development sweep, so the operating points of \method{} differ from those of the main text, which uses $\tau = 0.5$ on both tasks: 66.2\% at 31\% target-written steps instead of 66.6\% at 24\% on Mind2Web, and 50.7\% at 28\% instead of 51.0\% at 19\% on AndroidControl.}
\label{tab:s_judges_all}
\end{table}

\begin{figure}[!htb]
\centering
\includegraphics[width=\linewidth]{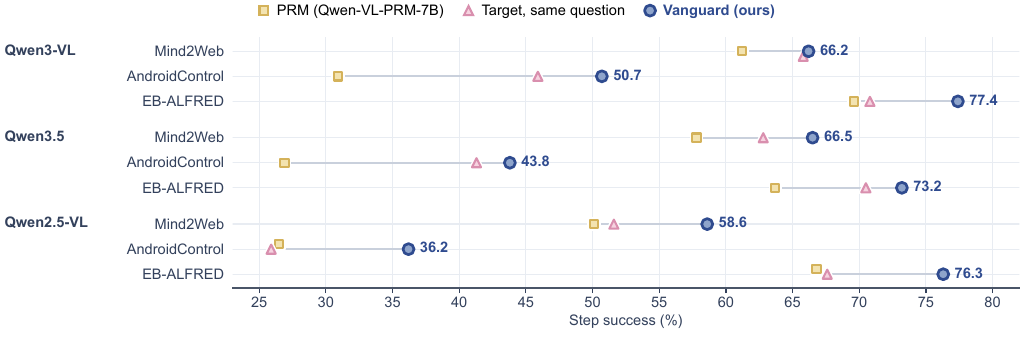}
\caption{\textbf{In all nine (family, benchmark) cells, \method{} is the most accurate of the three judges.} Higher is better. Step success (\%) of the pipeline with \method{} (navy, value printed), the process reward model Qwen-VL-PRM-7B (yellow square) or the target model asked the same question (pink triangle) as the judge. Full numbers: \cref{tab:s_judges_all}.}
\label{fig:s_judges_all}
\end{figure}

The comparison of judges in the main text uses one family and two benchmarks.
\Cref{tab:s_judges_all} extends it to the nine (family, benchmark) cells of the three families, adding the vision-language process reward model Qwen-VL-PRM-7B everywhere; every judge scores the same candidates of the draft, with the prompt format of the main text (for the PRM: its own system prompt, the screen, the task as the problem and the proposed action as the step).
\method{} is the most accurate judge in every cell (\cref{fig:s_judges_all}).
The PRM, trained on visual reasoning rather than on agent actions, falls below the target model alone in every cell (\cref{tab:judges,tab:s_alfred,tab:s_families} and \cref{sec:s_passk}); the target asked the same question is significantly behind in seven of the nine cells. In the other two it is also behind (65.8 against 66.2\% on Mind2Web with Qwen3-VL and 41.3 against 43.8\% on AndroidControl with Qwen3.5).

\paragraph{A Critic Trained for GUI Actions.}
Process reward models are trained on reasoning, not on agent actions.
GUI-Critic-R1~\cite{wanyan2025guicritic}, a Qwen2.5-VL-7B model trained to judge a GUI agent's next action before it is executed, is the closest specialised judge; we run it with its released weights and prompt (the screenshot with the action's position marked, the instruction, the history and the proposed action) and use its probability of ``Correct'' at the verdict token as the score.
On the same Qwen3-VL candidates, at the share of fallbacks of \method{}, it scores 57.0\% on Mind2Web and 42.5\% on AndroidControl, against 66.2\% and 50.7\% for \method{} (251/86 and 68/25 steps; $p \le 10^{-5}$), and its gate separates right from wrong candidates far less well (AUROC 0.583 and 0.654 against 0.807 and 0.714): its verdicts are almost always certain, so its best threshold on development data hands every step to the target.

\clearpage
\subsection{When Does the Draft Beat the Target?}
\label{sec:s_passk}

Selection alone cannot repair a step on which all four candidates are wrong, so the accuracy of the pipeline is bounded by the target's accuracy on the rejected steps and the draft's pass@4 on the accepted ones.
\Cref{fig:passk} shows sixteen (draft, target, task) triples, every result of a draft--target pair that this paper reports on Mind2Web, AndroidControl and EB-ALFRED: the main pair, the two families of \cref{tab:s_families} (also on AndroidControl), the Qwen3-VL-2B draft with the targets of \cref{sec:s_cross} and the Gemma-3-12B draft with the Gemma-3-27B target.
On AndroidControl (525 test steps, thresholds chosen on development data), the draft's pass@4, the target alone and \method{} reach 33.1, 37.1 and 36.2\% with Qwen2.5-VL 3B$\to$7B, whose taps, given in pixels of the image it sees, we convert to the 0--1000 scale of the prompt before scoring and judging, and 37.3, 47.4 and 43.8\% with Qwen3.5 2B$\to$9B; both triples lie within 0.7 points of the line fitted to the other fourteen.
Over all sixteen triples, the gain of \method{} over the target grows with the headroom, the draft's pass@4 (the share of steps on which at least one of its four samples is correct) minus the target's accuracy: the Pearson correlation is $r = 0.95$ and the Spearman correlation $\rho = 0.85$ (permutation tests with $10^6$ random orderings of the sixteen triples: $p < 10^{-5}$ and $p = 7\times10^{-5}$), and leaving out any one triple keeps $r \ge 0.92$.
The three largest gains (8.1, 8.4 and 11.8 points) come with the three largest headrooms (8.9, 15.3 and 18.7 points), and there \method{} turns 55--91\% of the headroom into accuracy.

\paragraph{Selection Among the Target Model's Own Samples.}
When the 8B samples four candidates itself and \method{} picks one without fallback, Mind2Web test\_task step success rises from 64.4\% to 71.1\% (71/16 steps, $p = 2\times10^{-9}$; an oracle over the four samples reaches 75.0\%), at about 1.36$\times$ the time of an 8B step.

\clearpage
\subsection{Predicting the Speedup}
\label{sec:s_cost}

\begin{table}[!htb]
\centering
\small
\setlength{\tabcolsep}{5pt}
\begin{tabular}{@{}lcc@{}}
\toprule
Time per step (s) & Mind2Web & AndroidControl \\
\midrule
Four draft candidates, $c_K$ & 1.26 & 1.57 \\
Decision model, $c_D$ & 0.56 & 0.42 \\
Target step, $c_p$ & 2.98 & 2.50 \\
\midrule
$c_K + c_D$ & 1.82 & 1.99 \\
$\alpha_\tau\, c_p$ & 2.26 & 2.02 \\
\bottomrule
\end{tabular}
\caption{\textbf{Drafting and judging cost less than the target steps that \method{} saves.} Measured components of the cost condition $c_K + c_D < \alpha_\tau c_p$ of the main text for \method{} at $\tau = 0.5$ (batch size 1, decision model through its public API). $\alpha_\tau$ is the share of accepted steps (76\% on Mind2Web, 81\% on AndroidControl); $c_D$ is the time added by judging the candidates.}
\label{tab:s_components}
\end{table}

\begin{table}[!htb]
\centering
\footnotesize
\setlength{\tabcolsep}{4pt}
\begin{tabular}{@{}lcc@{}}
\toprule
Setting (target 8B unless stated) & Predicted & Measured \\
\midrule
Mind2Web test\_task & 1.21 & 1.17 \\
Mind2Web test\_website & 1.19 & 1.15 \\
AndroidControl & 1.06 & 1.06 \\
Mind2Web test\_task, EAGLE-3 on the 8B & 1.32 & 1.28 \\
Mind2Web test\_task, EAGLE-3 on both & 1.54 & 1.50 \\
Mind2Web test\_website, EAGLE-3 on both & 1.55 & 1.52 \\
AndroidControl, EAGLE-3 on both & 1.57 & 1.58 \\
EB-ALFRED & 1.10 & 1.06 \\
EB-ALFRED, EAGLE-3 on the 8B & 1.21 & 1.19 \\
\bottomrule
\end{tabular}
\caption{\textbf{Three measured times and the fallback rate predict the measured speedup.} Mean speedup over the plain target model, predicted from per-component times as $c_p / (c_K + c_D + (1-\alpha_\tau)\,\tilde c_p)$, where $\tilde c_p$ is the time of a fallback step (equal to $c_p$ unless EAGLE-3 runs on the target), against the speedup measured end to end at batch size 1 on the same steps. The prediction uses the component times and fallback rate of each timed run itself, so it can differ slightly from the estimates of \cref{tab:s_components} and the full-sweep tables, which combine component times measured separately on other steps.}
\label{tab:s_cost}
\end{table}

\begin{figure}[!htb]
\centering
\includegraphics[width=\linewidth]{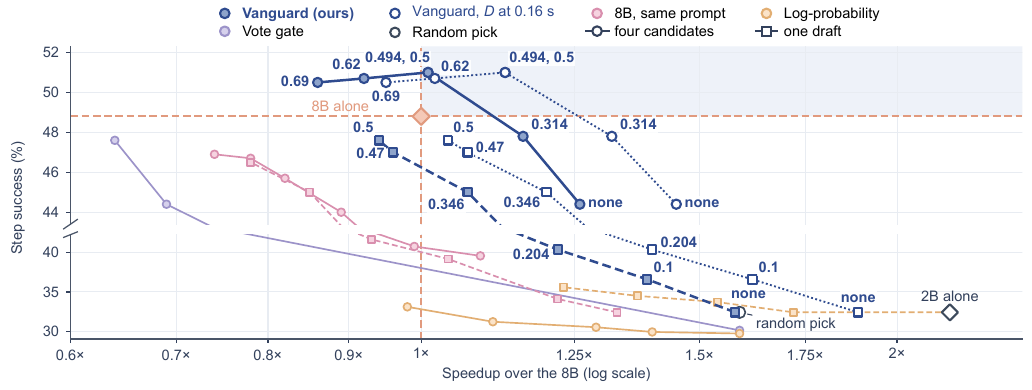}
\caption{\textbf{Only \method{} reaches the corner that is both faster and more accurate than the 8B alone.} Up and to the right is better. Step success against speedup over the 8B (log scale) on the AndroidControl test part (525 steps) for every configuration of the full threshold sweeps, coloured by judge. Circles and solid lines: four candidates; squares and dashed lines: one draft; each line joins the thresholds of one judge. Hollow navy markers on dotted lines: \method{} with the decision model at its nominal latency of 0.16\,s. Numbers next to \method{}'s points: $\tau$ (none: no fallback). Shaded: faster and more accurate than the 8B alone. The step-success axis is broken, with a larger scale above the break. Speedups are the per-component estimates of \cref{tab:s_android} over all 525 steps; the end-to-end measurement at $\tau = 0.5$ is 1.06$\times$ (\cref{tab:s_cost}). Full numbers: \cref{tab:s_android}.}
\label{fig:s_speed_android}
\end{figure}

\begin{figure}[!htb]
\centering
\includegraphics[width=0.5\linewidth]{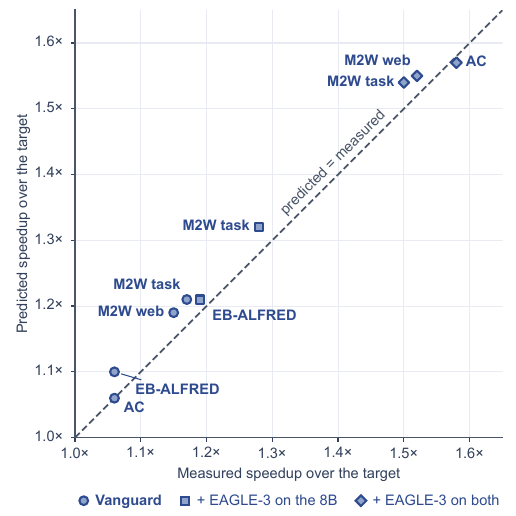}
\caption{\textbf{In all nine settings, the predicted speedup is within 0.04$\times$ of the measured one.} Speedup over the plain target model predicted from the component times and the fallback rate, against the measured speedup; dashed line: predicted = measured. Circles: \method{}; squares: with EAGLE-3 on the 8B; diamonds: with EAGLE-3 on both the draft and the target. M2W task / web: Mind2Web test\_task / test\_website; AC: AndroidControl. Full numbers: \cref{tab:s_cost}.}
\label{fig:s_speed_pred}
\end{figure}

The condition of the main text also predicts how much time \method{} saves.
\Cref{tab:s_cost} compares, for the settings timed end to end, the speedup predicted from the per-component times and the measured share of target-written steps with the measured one: the two agree within 0.04$\times$ in all nine (\cref{fig:s_speed_pred}).
Three measurements per setting, the cost of four draft samples, of the judge and of a target step, together with the fallback rate, thus suffice to predict the speedup of a configuration.
\Cref{tab:s_components} gives the components for the main configuration: drafting four candidates and judging them takes less time than the target steps that \method{} saves, and \cref{fig:s_speed_android} places every AndroidControl configuration by speed and accuracy.

Of all configurations in the sweeps, only those of \method{} lie in the shaded corner, faster and more accurate than the 8B alone (\cref{fig:s_speed_android}).

\clearpage
\subsection{The Composed Configuration on the Full Test Sets}
\label{sec:s_compose_full}

The speed measurements of \cref{tab:speed} (\cref{fig:speed}) ran the composed configuration (EAGLE-3 on both the draft and the target) on subsets of 300 steps.
Run on the full Mind2Web test sets with the same thresholds, it scores 66.2\% on test\_task (820 steps) and 68.4\% on test\_website (975), against 65.1 and 63.5 for the 8B with EAGLE-3 alone, 64.4 and 63.3 for the plain 8B, and 66.6 (at the threshold 0.41 of \cref{tab:speed}) and 66.8 for \method{} without EAGLE-3.
The composed configuration is significantly more accurate than EAGLE-3 alone on test\_website (111/63 steps, $p = 3\times10^{-4}$). It is also ahead of EAGLE-3 alone on test\_task (66.2 against 65.1\%) and on both timed subsets of \cref{tab:speed} (70.0 against 67.0 and 62.0 against 59.7\%), and so is \method{} without EAGLE-3 in all four measurements; what the composition adds on top of \method{} is speed (1.50 and 1.52$\times$ on the timed subsets, against 1.17 and 1.15$\times$ without EAGLE-3).

\paragraph{Training-Free Token-Level Speculation.}
Prompt-lookup (n-gram) speculative decoding in vLLM also needs no trained component; on the first 200 steps of test\_website it makes the 8B only 1.02$\times$ faster (3.00 against 3.08\,s per step), against 1.57$\times$ for EAGLE-3 on the same steps.

\clearpage
\subsection{Draft-Model Speculative Decoding}
\label{sec:s_short_drafts}

Classic speculative decoding with a small draft model is not available for these vision-language models in vLLM, which supports only EAGLE-style heads and n-gram drafts for them.
With Hugging Face Transformers' assisted generation, where Qwen3-VL-2B drafts tokens for Qwen3-VL-8B, decoding is slower than the 8B alone on 50 Mind2Web steps (0.58$\times$ on average: 7.22\,s against 4.19\,s per step), because every draft round re-runs the draft model on the full multimodal context; EAGLE-3, the token-level baseline of the main text, is the strongest token-level option we could run.

\clearpage
\subsection{Number of Candidates, Sampling Seeds and Temperature}
\label{sec:s_seeds}

\begin{figure}[!htb]
\centering
\includegraphics[width=0.55\linewidth]{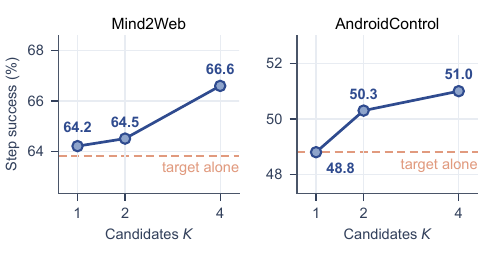}
\caption{\textbf{More candidates raise accuracy on Mind2Web while fewer steps go to the target.} Higher is better. Step success of \method{} with the first $K$ of the four draft candidates. On Mind2Web, the share of target-written steps falls from 40\% at $K = 1$ to 31\% and 24\% at $K = 2$ and $4$.}
\label{fig:s_k}
\end{figure}

With the first $K$ of the four candidates (\cref{fig:s_k}), accuracy rises with $K$ on both tasks, and on Mind2Web the share of target-written steps falls.

\begin{table}[!htb]
\centering
\footnotesize
\setlength{\tabcolsep}{1.8pt}
\begin{tabular}{@{}lccccc@{}}
\toprule
& & \multicolumn{3}{c}{\method{}, $K = 4$} & $K = 8$ \\
\cmidrule(lr){3-5}\cmidrule(l){6-6}
Test set & Target & Seed 0 & Seed 1 & Seed 2 & Success (Target) \\
\midrule
Mind2Web test\_task & 64.4 & 66.5 & 66.3 & 66.6 & 68.3 (15) \\
Mind2Web test\_website & 63.3 & 66.8 & 67.2 & 67.1 & 68.6 (18) \\
AndroidControl & 48.8 & 51.0 & 49.1 & 49.3 & 50.7 (12) \\
\bottomrule
\end{tabular}
\caption{\textbf{\method{} stays above the target with every draw of the draft's samples, and eight candidates raise it further on Mind2Web.} Step success (\%) of \method{} at $\tau = 0.5$ ($\tau = 0.41$ on test\_task, the threshold of the speed measurements; in this table a tie in $D$ goes to the first candidate) when the draft model's four samples are redrawn with another seed, and with eight candidates per step (the samples of seeds 0 and 1 pooled; share of target-written steps, \%, in parentheses). With $K = 8$, \method{} beats the 8B on 80/48 steps on test\_task ($p = 0.006$) and 111/59 on test\_website ($p = 8\times10^{-5}$); the oracle over the candidates rises from 62.3 to 69.8 and from 63.3 to 70.5.}
\label{tab:s_seeds}
\end{table}

\begin{figure}[!htb]
\centering
\includegraphics[width=0.6\linewidth]{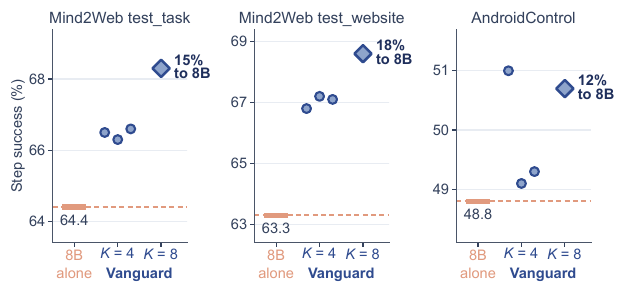}
\caption{\textbf{With every draw of the draft's samples and with eight candidates, \method{} stays above the target alone on all three test sets.} Higher is better. Step success (\%) of the 8B target alone (peach tick and dashed line) and of \method{} with $K = 4$ candidates for sampling seeds 0, 1 and 2 (dots, left to right) and with $K = 8$ (diamond, labelled with the share of target-written steps). On both Mind2Web test sets, eight candidates are above every seed of $K = 4$. Each panel has its own y-axis range. Full numbers: \cref{tab:s_seeds}.}
\label{fig:s_seeds}
\end{figure}

Redrawing the draft model's four samples with another seed moves the step success of \method{} by a few tenths of a point on Mind2Web (standard deviation over three seeds 0.1 and 0.2 points) and by about a point on AndroidControl (1.0 point; \cref{tab:s_seeds}), where the test set is smaller (525 steps). Every result of the main text uses seed 0, the first draw; seeds 1 and 2 were drawn afterwards. Averaged over the three seeds, \method{} reaches 49.8\% on AndroidControl against 48.8\% for the target alone.
Eight candidates per step raise the ceiling further: the oracle over the candidates gains about seven points on both Mind2Web test sets, and on Mind2Web \method{} gains about two points over $K = 4$ while sending fewer steps to the target, at the cost of judging twice as many candidates; on AndroidControl it stays at the level of $K = 4$. There, accuracy rises from the level of the 8B alone with one candidate (48.8\%) to 50.3 and 51.0\% with two and four (\cref{fig:s_k}), and eight candidates keep it at 50.7\% while lowering the share of target-written steps from 19 to 12\%.
With vLLM the four samples come from one batched request at nearly the cost of one, so drawing all four at once is also the fastest choice: a second request costs more than the two samples it saves, and in our offline estimate every rule that draws two candidates first and two more only when the judge is not confident is slower than $K = 4$ (0.86$\times$ against 1.20$\times$ on test\_task) at the same accuracy.
The sampling temperature of the draft model matters little: at $T = 0.3$, $0.7$ (our default) and $1.0$, \method{} scores 66.9, 66.6 and 66.4\% on the pooled Mind2Web test sets (all within half a point), with 25, 22 and 21\% of the steps written by the 8B (on test\_task the sweep uses $\tau = 0.41$, the threshold of the speed measurements, so the share at the default differs slightly from \cref{tab:m2w}); a higher temperature gives more diverse candidates (1.61, 1.86 and 2.06 distinct actions among the four) and a slightly higher pass@4 (59.3, 62.8 and 63.1\%), which the judge turns into slightly fewer fallbacks rather than a higher accuracy. \Cref{fig:s_seeds} shows the seeds and $K = 8$ against the target.

\clearpage
\subsection{What the Judge Needs to See}
\label{sec:s_inputs}

\begin{table}[!htb]
\centering
\footnotesize
\setlength{\tabcolsep}{2pt}
\begin{tabular}{@{}lcccc@{}}
\toprule
Input of $D$ & Success $\uparrow$ & vs.\ \method{} & AUROC $\uparrow$ \\
\midrule
\rowcolor{vgmist}\method{} (screen, history, action) & \textbf{66.6} & -- & 0.807 \\
No screenshot & 63.1 & 45/108 & 0.764 \\
Screenshot of another task & 60.8 & 57/161 & 0.673 \\
No previous actions & 64.6 & 17/54 & 0.766 \\
With the draft's reasoning & 62.5 & 45/119 & 0.756 \\
\midrule
\multicolumn{4}{@{}l}{\emph{Wording of the question}} \\
Negated question, $1 - P(\text{yes})$ & 64.6 & 35/72 & 0.794 \\
\bottomrule
\end{tabular}
\caption{\textbf{The judge needs the screenshot and the history, and only the action.} Mind2Web, both test sets, 1,795 steps. Each variant rescores the same candidates with one input or the question changed; its threshold is chosen on development data to send the same share of steps to the 8B as \method{} (22--24\% on test). ``vs.\ \method{}'' gives steps won/lost against \method{}; removing or replacing any input, adding the reasoning and negating the question are all significantly worse ($p \le 5\times10^{-4}$). AUROC is that of the gate on the selected candidate. Negated question: ``Is this action a mistake?''}
\label{tab:s_inputs}
\end{table}

\begin{figure}[!htb]
\centering
\includegraphics[width=0.6\linewidth]{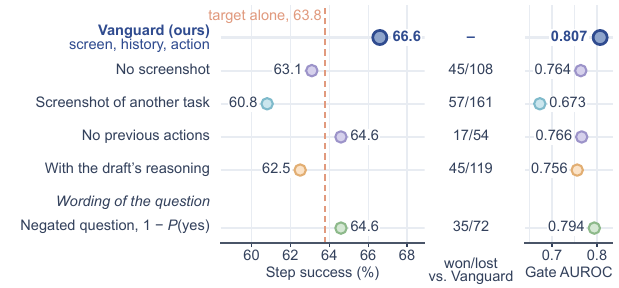}
\caption{\textbf{The judge is most accurate and best at separating right from wrong candidates when it sees the screen, the history and only the action.} Higher step success and higher AUROC are better. Mind2Web, both test sets (1,795 steps). Left: step success (\%) when one input of $D$ is removed, replaced or added, or the question is negated; dashed line: the 8B target alone. Middle: steps won/lost against \method{}. Right: AUROC of the gate on the selected candidate. Full numbers: \cref{tab:s_inputs}.}
\label{fig:s_inputs}
\end{figure}

\Cref{tab:s_inputs,fig:s_inputs} change one input of the decision model at a time on Mind2Web, with everything else, including the share of target-written steps, held fixed.
Both the screenshot and the previous actions are needed: without either, \method{} falls back to the level of the 8B alone.
Showing the decision model the draft's reasoning as well as its action makes it significantly worse (62.5\% against 66.6\%): a fluent justification makes wrong actions more convincing, so the judge should check the action against the screen rather than read the argument for it, unlike schemes in which the target model grades each reasoning step.
The polarity of the question matters: asking whether the action is a mistake and inverting the probability is significantly worse.

On EB-ALFRED, \method{} with the frame also remains the most accurate configuration (77.4\% against 76.7\% without the frame and 76.5\% with the frame of another episode, at the same share of fallbacks; 53/66 and 54/71 steps won/lost against \method{}), with the highest gate AUROC (0.859 against 0.849 and 0.810).
On MathVista, where a step is a paragraph of reasoning about a figure, the figure is needed as well.
Running the speculative pipeline on 799 of the 800 held-out problems with the decision model shown the problem and the reasoning but not the figure, with thresholds chosen on the development problems to give the same share of target-written steps (1.8\% and 10.0\% on development), lowers accuracy from 68.3\% to 62.7\% (71/26 problems, $p = 5\times10^{-6}$) and from 69.7\% to 65.0\% (69/31, $p = 2\times10^{-4}$); the reference with the figure is the pipeline of \cref{tab:s_mathvista} on the same problems, at thresholds matched on development data ($\tau = 0.24$ and $0.49$; 3.9\% and 11.5\% target-written steps on test); these are not the operating points shown in \cref{tab:s_mathvista}.
With the figure of another problem, the decision model instead rejects most steps (8\% of the steps go to the 8B on development already at $\tau = 0.05$), so the pipeline falls back to the target rather than accepting wrong steps.

\clearpage
\subsection{Where the Gain Comes From}
\label{sec:s_where}

\begin{table}[!htb]
\centering
\footnotesize
\setlength{\tabcolsep}{1.5pt}
\begin{tabular}{@{}lccc@{}}
\toprule
& Mind2Web & AndroidControl & EB-ALFRED \\
\midrule
Steps accepted & 76 & 81 & 72 \\
Accepted: draft / target right & 71.6 / 67.9 & 51.3 / 48.5 & 86.5 / 82.1 \\
Rejected: draft / target right & 12.7 / 50.9 & 15.0 / 50.0 & 30.2 / 53.7 \\
\midrule
Fallbacks & 432 & 100 & 514 \\
\quad target fixes a wrong draft & 189 (71) & 40 (14) & 169 (67) \\
\quad target replaces a right draft & 55 (248) & 15 (45) & 155 (364) \\
\bottomrule
\end{tabular}
\caption{\textbf{The gate keeps the drafts that beat the target and sends it the steps it can fix.} Decisions of \method{} on the test steps (Qwen3-VL 2B $\to$ 8B). Percentages except for the counts of fallbacks; ``draft'' is the candidate selected by \method{} on that step, ``target'' the 8B's own action on the same step. In parentheses: the expected counts for a gate that falls back at random at the same rate.}
\label{tab:s_where}
\end{table}

\begin{figure}[!htb]
\centering
\includegraphics[width=0.55\linewidth]{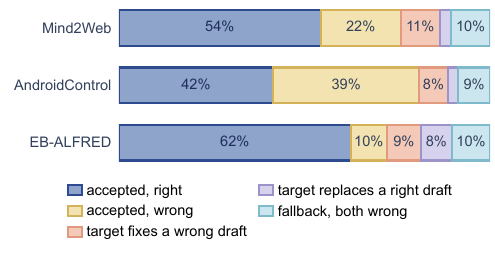}
\caption{\textbf{Accepted drafts that are right form the largest group on every task.} Where each test step ends up under \method{}: accepted (right or wrong), or handed to the 8B, which fixes a wrong draft, rejects a draft that was right, or is wrong as well.}
\label{fig:s_outcomes}
\end{figure}

\begin{figure}[!htb]
\centering
\includegraphics[width=\linewidth]{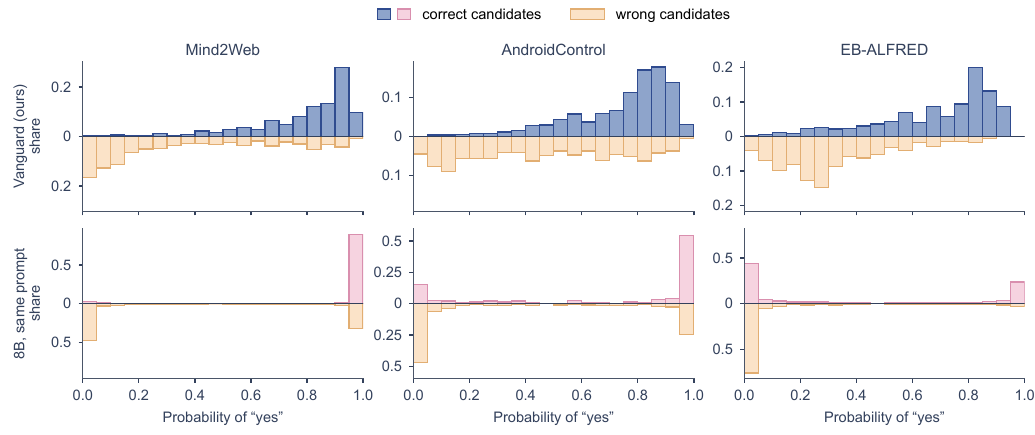}
\caption{\textbf{\method{} spreads its scores and separates right from wrong candidates, whereas the 8B's scores pile up at 0 and 1.} Scores of correct and wrong candidates on the test steps. Top: \method{}; bottom: the 8B asked the same question. The 8B's probabilities are almost always near 0 or 1, so a threshold on them either accepts or rejects nearly everything.}
\label{fig:s_hist}
\end{figure}

Since a rejected step is written by the target, every step won or lost against the target is an accepted one.
On the accepted steps, the candidate selected by \method{} is right more often than the target's own action on the same steps, and on the rejected steps it would have been right only 13--30\% of the time (\cref{tab:s_where}, \cref{fig:s_outcomes}).
The gate thus sends the target the steps it can fix: on Mind2Web, 189 of the 432 fallbacks replace a wrong draft by a right action, against 71 expected for a random gate at the same rate, and only 55 discard a draft that was right, against 248.
The scores of \method{} separate correct from wrong candidates over the whole range (\cref{fig:s_hist}); the selected draft is right 84\% of the time above 0.9 on Mind2Web and 99\% between 0.9 and 0.95 on EB-ALFRED, whereas the target asked the same question puts 77\% of the Mind2Web steps above 0.95.

By the share of steps, accepted drafts that are right are the largest group on every task (\cref{fig:s_outcomes}).

\FloatBarrier
\FloatBarrier

By action type, the gain on Mind2Web comes from clicks (140 steps won, 62 lost; 70.6\% against 65.2\%), the most frequent action type, where the target's errors are mostly a wrong element.
On AndroidControl, the gain is again in clicks (33/23); on EB-ALFRED it comes from both manipulation (92/54) and navigation (40/19).
Four candidates often agree: all four propose the same action on 40\% of the Mind2Web steps, 8\% of the AndroidControl steps and 63\% of the EB-ALFRED steps; even then, where only the gate acts, \method{} is 1.8 points above the target on Mind2Web.

\clearpage
\subsection{Reasoning Benchmarks}
\label{sec:s_reasoning}
\begin{table}[!htb]
\centering
\small
\setlength{\tabcolsep}{4pt}
\begin{tabular}{@{}lcc@{}}
\toprule
& \multicolumn{2}{c}{Steps written by the target model} \\
\cmidrule(l){2-3}
Judge & $\approx$2\% & $\approx$12\% \\
\midrule
VisualPRM-8B & 63.2 (2.5) & 63.4 (13.4) \\
Qwen3-VL-8B, same prompt & 64.5 (2.1) & 66.7 (12.0) \\
Qwen-VL-PRM-7B & 67.6 (2.5) & 69.5 (15.1) \\
\rowcolor{vgmist}\method{} & \textbf{67.9} (1.5) & 69.4 (11.9) \\
\midrule
\multicolumn{3}{@{}l}{Draft alone 58.5 \quad Random of four 60.4} \\
\bottomrule
\end{tabular}
\caption{\textbf{On MathVista, \method{} is on par with Qwen-VL-PRM-7B and ahead of VisualPRM-8B and the 8B asked the same question.} Accuracy (\%) on 800 held-out testmini problems with four candidates per step and 8B fallback; the exact share of target-written steps (\%) is given in parentheses. With Holm correction over the six comparisons of this table, \method{} is significantly more accurate than VisualPRM at both shares ($p \le 0.004$) and than the 8B asked the same question at the lower share ($p = 0.020$).}
\label{tab:s_mathvista}
\end{table}

\Cref{tab:s_mathvista,tab:s_text} give the MathVista and text-reasoning results summarised in the main text; on MathVista, \method{} is on par with Qwen-VL-PRM-7B at both shares of target-written steps. Against the 8B asked the same question, \method{} is ahead at both shares (67.9 against 64.5\% and 69.4 against 66.7\%), significantly so at the lower share after Holm correction over the six comparisons of \cref{tab:s_mathvista}.

\begin{table}[!htb]
\centering
\small
\setlength{\tabcolsep}{4pt}
\begin{tabular}{@{}lcc@{}}
\toprule
ProcessBench & \multicolumn{2}{c}{F1, held-out subsets} \\
\midrule
Qwen3-8B, same prompt & \multicolumn{2}{c}{49.9} \\
Qwen3-32B, same prompt & \multicolumn{2}{c}{61.0} \\
Qwen2.5-Math-PRM-7B & \multicolumn{2}{c}{63.0} \\
\rowcolor{vgmist}\method{} & \multicolumn{2}{c}{\textbf{66.6}} \\
\midrule
MATH-500 test, best of four & \multicolumn{2}{c}{Accuracy} \\
\midrule
Qwen3-8B as judge & \multicolumn{2}{c}{79.0} \\
Qwen2.5-Math-PRM-7B & \multicolumn{2}{c}{79.7} \\
\rowcolor{vgmist}\method{} & \multicolumn{2}{c}{\textbf{82.8}} \\
\bottomrule
\end{tabular}
\caption{\textbf{On text reasoning, \method{} is ahead of the general models asked the same question and of the trained math process reward model.} Top: ProcessBench F1 for locating the first erroneous step, averaged over MATH, OlympiadBench and Omni-MATH, with every threshold chosen on GSM8K and every general model asked the \emph{correct} question. Bottom: accuracy (\%) on the 400 test problems of MATH-500 when the judge picks one of four complete draft solutions (draft alone 73.8); \method{} beats the math process reward model on 15 problems against 3 ($p = 0.008$).}
\label{tab:s_text}
\end{table}

In \cref{tab:s_text}, \method{} is ahead of the general models asked the same question and of the trained math process reward model, which it beats on 15 MATH-500 problems against 3.

\begin{table}[!htb]
\centering
\footnotesize
\setlength{\tabcolsep}{6pt}
\begin{tabular}{@{}lcccccc@{}}
\toprule
Judge & $\tau$ & GSM8K & MATH & Olymp. & Omni & Held-out \\
\midrule
Qwen3-8B, same (correct) & 0.08 & 55.8 & 56.2 & 46.9 & 46.6 & 49.9 \\
Qwen3-8B, same (compute) & 0.51 & 56.2 & 54.2 & 40.8 & 35.8 & 43.6 \\
Qwen3-8B, same (error) & 0.98 & 52.4 & 37.1 & 36.4 & 37.5 & 37.0 \\
Qwen3-32B, same (correct) & 0.51 & 74.4 & 67.2 & 59.0 & 56.9 & 61.0 \\
Qwen3-32B, same (compute) & 0.57 & 73.9 & 64.0 & 54.0 & 53.5 & 57.2 \\
Qwen3-32B, same (error) & 0.05 & 71.8 & 52.9 & 34.1 & 29.7 & 38.9 \\
\rowcolor{vgmist}\method{} (correct) & 0.58 & 74.7 & 71.6 & 62.6 & 65.6 & 66.6 \\
\rowcolor{vgmist}\method{} (compute) & 0.53 & 58.9 & 70.8 & 64.6 & 65.8 & 67.1 \\
\method{} (error) & 0.83 & 77.1 & 65.5 & 52.3 & 52.2 & 56.7 \\
\bottomrule
\end{tabular}
\caption{\textbf{With either text-reasoning question, \method{} is ahead of every general model asked the same question on the held-out subsets.} ProcessBench F1 per subset for every check question we tried (\emph{correct}, \emph{compute} and a third question, \emph{error}); \cref{tab:s_text} reports the \emph{correct} question for every general model and, for \method{}, the better of the two text-reasoning questions of the main text (\emph{correct} and \emph{compute}) on GSM8K, where thresholds are chosen; with either of these two questions, \method{} is ahead of every other judge whose threshold is chosen on GSM8K on the held-out average. ``Held-out'' averages MATH, OlympiadBench and Omni-MATH.}
\label{tab:s_pb}
\end{table}

Per subset (\cref{tab:s_pb}), with either text-reasoning question \method{} is ahead of every general model asked the same question on the held-out subsets.

\begin{table}[!htb]
\centering
\footnotesize
\setlength{\tabcolsep}{3pt}
\begin{tabular}{@{}lccc@{}}
\toprule
Method & Acc. & Median & Mean \\
\midrule
8B alone & 78 & 1.0$\times$ & 1.0$\times$ \\
\method{} ($\tau = 0.7$) & 80 & 2.0$\times$ & 1.2$\times$ \\
\bottomrule
\end{tabular}
\caption{\textbf{On MATH-500, \method{} matches the accuracy of the 8B and is faster.} Accuracy (\%) and measured speedups on 50 MATH-500 development problems, one request at a time.}
\label{tab:s_latency}
\end{table}

On MATH-500, \method{} matches the accuracy of the 8B and is faster (\cref{tab:s_latency}).

\FloatBarrier

\clearpage
\subsection{Paired Tests for the Main-Text Comparisons}
\label{sec:s_pairs}

\Cref{tab:s_pairs} gives the exact McNemar tests for the remaining comparisons of the main figures and tables in which \method{} is significantly ahead, at the operating points shown there.

\begin{table}[!htb]
\centering
\footnotesize
\setlength{\tabcolsep}{5pt}
\begin{tabular}{@{}llcrr@{}}
\toprule
Task (main-text item) & \method{} against & $n$ & W/L & $p$ \\
\midrule
Mind2Web test\_task (\cref{tab:m2w}) & Qwen-VL-PRM-7B & 820 & 70/42 & 0.010 \\
 & Vote gate & 820 & 117/32 & $1\times10^{-12}$ \\
 & Random of four & 820 & 201/25 & $2\times10^{-35}$ \\
 & Draft model alone & 820 & 224/27 & $8\times10^{-40}$ \\
Mind2Web test\_website (\cref{tab:m2w}) & Qwen-VL-PRM-7B & 975 & 104/48 & $6.5\times10^{-6}$ \\
 & Vote gate & 975 & 133/37 & $6.5\times10^{-14}$ \\
 & Random of four & 975 & 268/32 & $1\times10^{-47}$ \\
 & Draft model alone & 975 & 256/32 & $1.5\times10^{-44}$ \\
Mind2Web, both test sets (\cref{fig:share}) & \makecell[l]{8B, same question ($\tau = 0.5$),\\ \method{} at $\tau = 0.41$} & 1,795 & 99/60 & 0.0025 \\
Mind2Web, both test sets (\cref{sec:exp_ablation}) & $K = 1$ & 1,795 & 107/63 & $9\times10^{-4}$ \\
 & $K = 2$ & 1,795 & 67/29 & $1\times10^{-4}$ \\
Mind2Web, timed steps (\cref{fig:speed}) & 8B alone, test\_task ($\tau = 0.41$) & 300 & 32/15 & 0.019 \\
 & 8B alone, test\_website & 300 & 30/12 & 0.008 \\
AndroidControl (\cref{fig:agents}) & 8B, same question (46.9\%) & 525 & 39/17 & 0.005 \\
 & Draft model alone & 525 & 119/21 & $8\times10^{-18}$ \\
AndroidControl (\cref{fig:share}) & \makecell[l]{8B, same question (46.7\%),\\ \method{} at $\tau = 0.62$} & 525 & 37/16 & 0.0055 \\
 & \makecell[l]{8B, same question (46.9\%),\\ \method{} at $\tau = 0.69$} & 525 & 34/15 & 0.009 \\
 & Vote gate (44.4\%, 82\% target-written) & 525 & 64/29 & $4\times10^{-4}$ \\
AndroidControl (\cref{fig:judges}) & 2B, self-check & 525 & 69/20 & $2\times10^{-7}$ \\
 & 8B, 0--9 grade & 525 & 49/13 & $5\times10^{-6}$ \\
EB-ALFRED (\cref{fig:agents}) & 8B, same question & 1,849 & 112/63 & $3\times10^{-4}$ \\
 & Draft model alone & 1,849 & 388/77 & $6\times10^{-51}$ \\
GoToLocal (\cref{fig:agents}) & 8B, same question & 100 & 25/2 & $6\times10^{-6}$ \\
 & Draft model alone & 100 & 49/0 & $4\times10^{-15}$ \\
PickupLoc (\cref{fig:agents}) & Draft model alone & 100 & 53/4 & $6\times10^{-12}$ \\
\bottomrule
\end{tabular}
\caption{\textbf{Every comparison of the main text tested here is significant.} Exact two-sided McNemar tests for the remaining comparisons of the main text (W/L: steps or episodes on which \method{} is right and the other method wrong, and the reverse). \method{} at $\tau = 0.5$ on the agent tasks ($\tau = 0.451$ on EB-ALFRED, $0.3$ in closed loop) unless stated otherwise; every other method at the operating point drawn in the main text (in parentheses: its step success, \%).}
\label{tab:s_pairs}
\end{table}
\FloatBarrier

\clearpage
\section{Qualitative Examples}
\label{sec:s_qualitative}

\FloatBarrier
\subsection{Case Studies}
\label{sec:s_cases}

\begin{figure}[!t]
\centering
\includegraphics[width=0.95\linewidth]{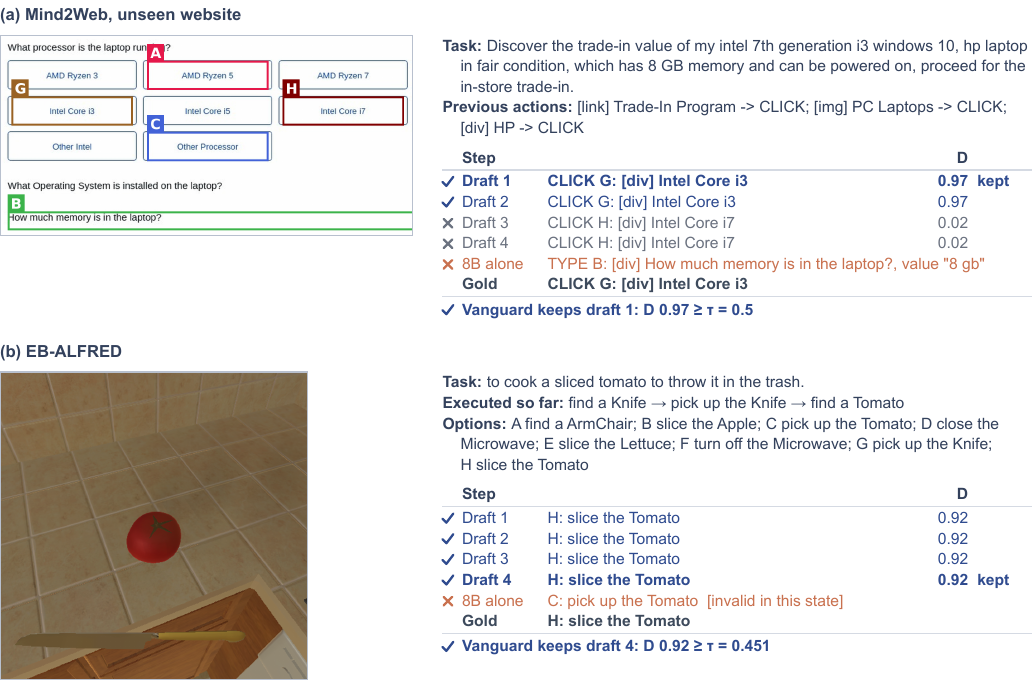}
\caption{\textbf{$D$ keeps a right draft on steps where the 8B alone is wrong.} (a) Mind2Web, unseen website: the step selects the processor; two drafts choose the right one (G, Intel Core i3) and two a wrong one (H), which $D$ scores 0.97 and 0.02, while the 8B skips ahead to the memory field (B); letters: Set-of-Mark tags as shown to the models. (b) EB-ALFRED: the robot holds a knife and has found the tomato; the 8B tries to pick the tomato up, which is impossible while holding the knife, whereas all four drafts slice it ($D$ 0.92, $\tau=0.451$).}
\label{fig:s_cases_win}
\vspace{4pt}
\includegraphics[width=0.95\linewidth]{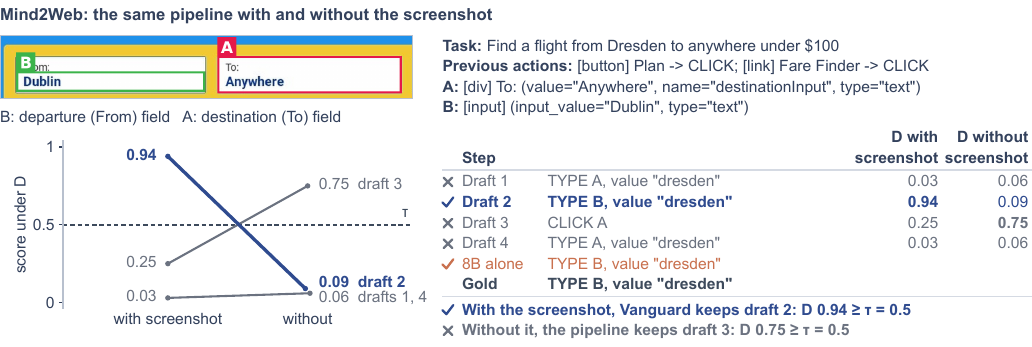}
\caption{\textbf{With the screenshot, $D$ keeps the draft that types into the right field.} The same pipeline with and without the screenshot on a Mind2Web step: with it, $D$ separates typing the city into the departure field B (0.94, right) from the destination field A (0.03); without it, it accepts a click on the destination field (0.75, above $\tau=0.5$).}
\label{fig:s_cases_fail}
\end{figure}

\begin{figure}[!t]
\centering
\includegraphics[width=\linewidth]{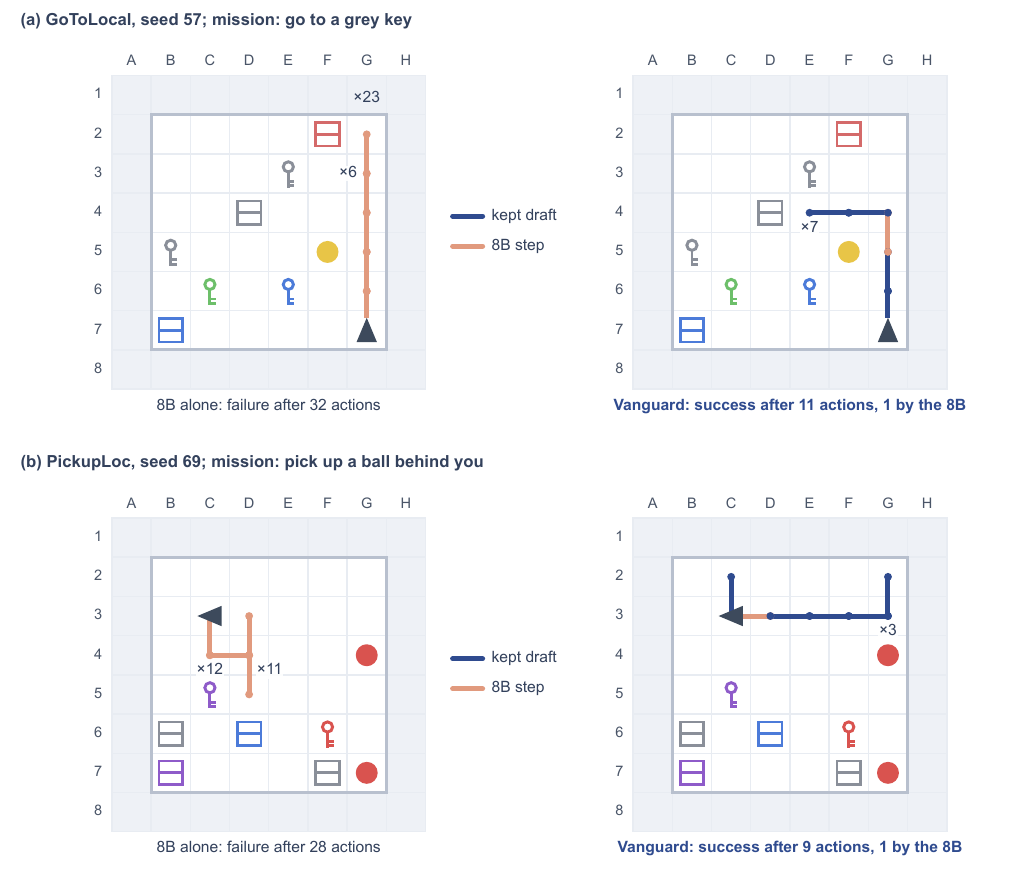}
\caption{\textbf{In closed loop, \method{} reaches the goal where the 8B alone gets stuck.} Closed-loop episodes on the same seeds, with each method's path drawn on the start frame (navy: steps kept from the 2B draft; peach: steps written by the 8B, for \method{} after a rejection; $\times n$: number of actions taken in that cell). The 8B alone gets stuck and runs out of steps; \method{} reaches the goal in 11 and 9 actions, with one decision each written by the 8B.}
\label{fig:s_cases_loop}
\end{figure}

\begin{figure}[!t]
\centering
\includegraphics[width=0.86\linewidth]{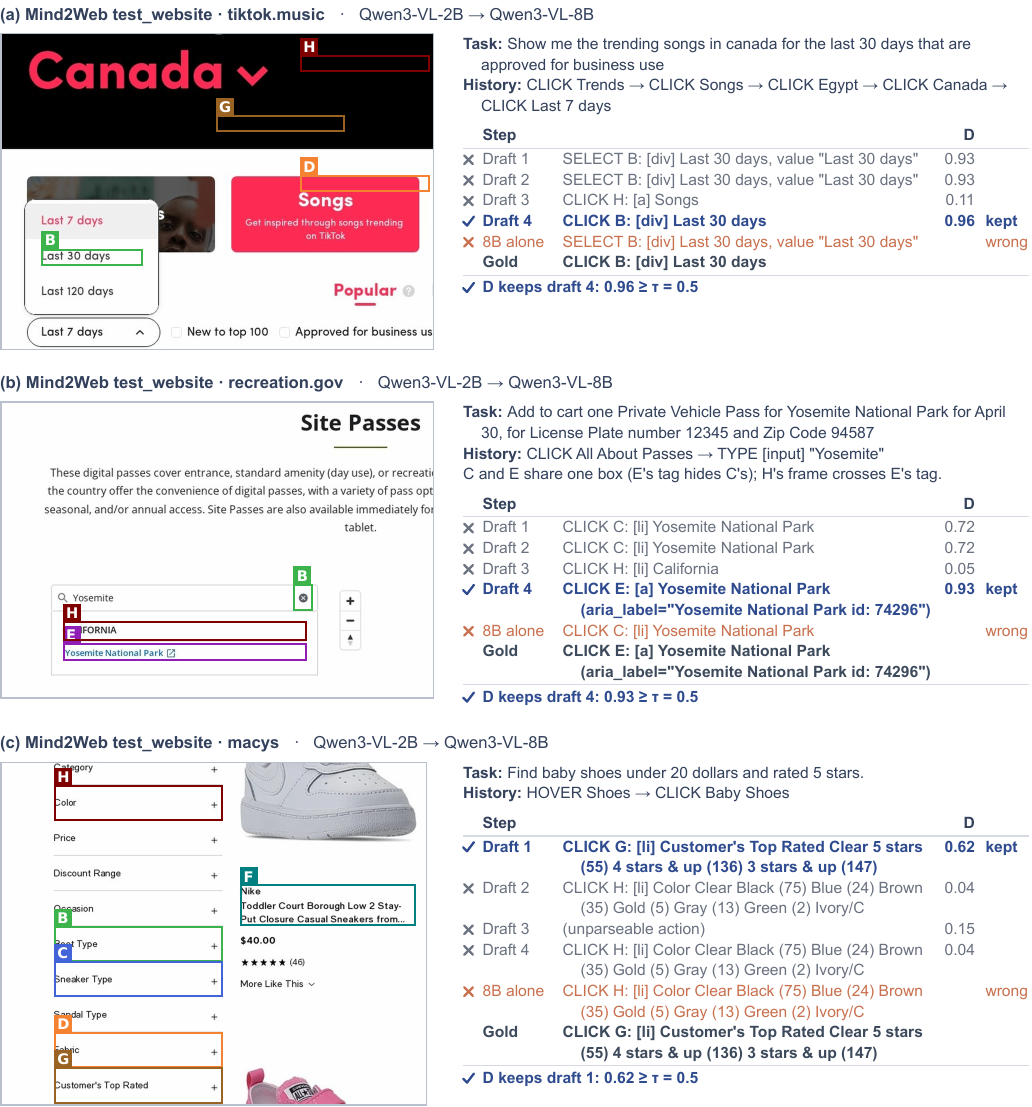}
\caption{\textbf{On unseen websites, $D$ picks out the right draft even where the 8B target alone goes wrong.} Mind2Web test\_website, draft Qwen3-VL-2B, target Qwen3-VL-8B, $\tau=0.5$. A step is shown if the four drafts propose at least two different actions (drafts with the same element, operation and value count as one; unparseable drafts are not counted), \method{} keeps a right draft, and the target alone is wrong; we show the first such step of each of the first 3 qualifying tasks in test order, out of 47 qualifying tasks (65 steps; the task of \cref{fig:s_cases_win} is not counted). Left: the model's input view at native resolution with its Set-of-Mark boxes, cropped around the chosen elements. Right: the four drafts with their $D$ scores (tick: right, cross: wrong), the 8B target alone, the gold action and $D$'s decision.}
\label{fig:s_gallery_m2w}
\end{figure}

\begin{figure}[!t]
\centering
\includegraphics[width=0.86\linewidth]{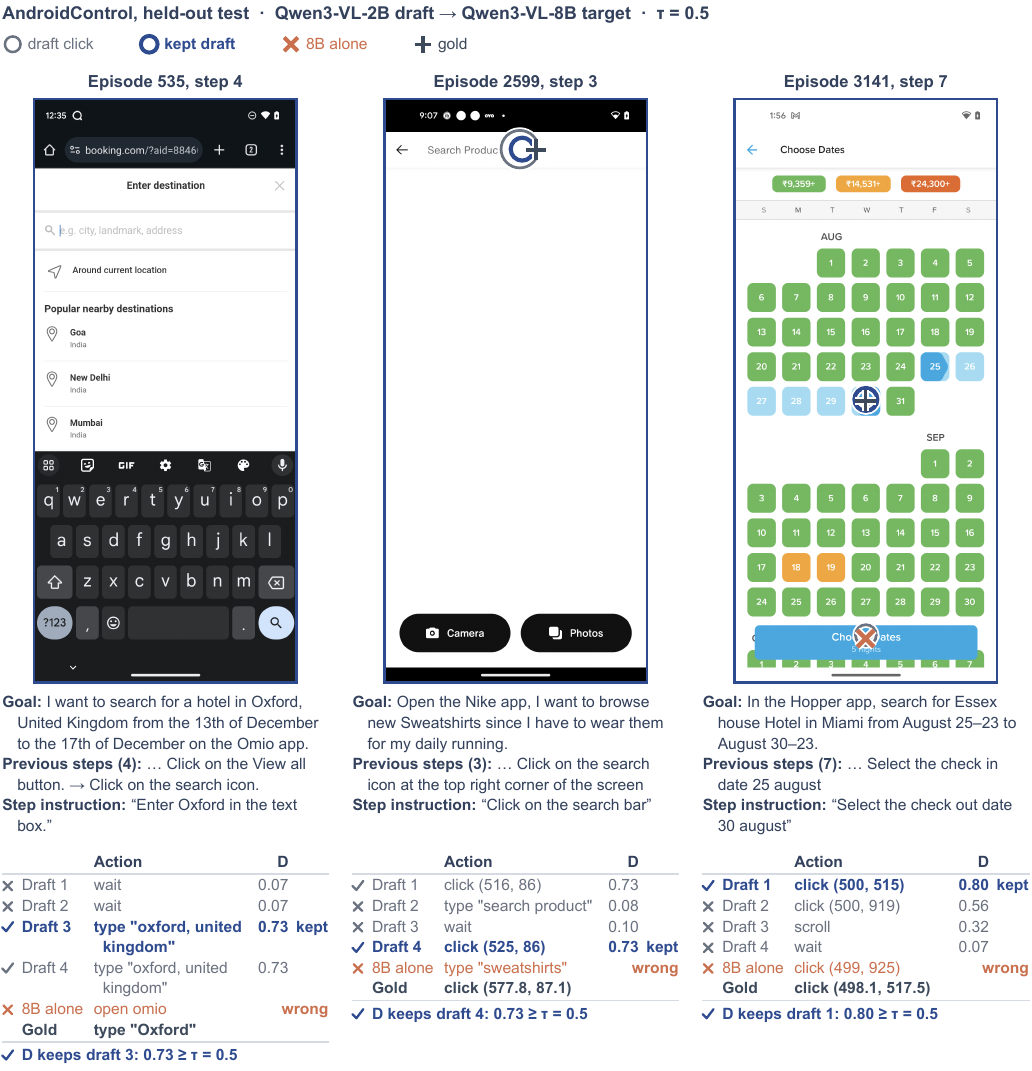}
\caption{\textbf{On held-out AndroidControl steps where the 8B alone is wrong, $D$ singles out a right 2B draft among the drafts' different proposals, and \method{} keeps it.} We select steps by a fixed rule: the four drafts propose at least two different actions (clicks within the success radius count as one), \method{} keeps a right draft, and the target alone is wrong; we show the first such step of each of the first 3 qualifying episodes in test order, out of 20 qualifying episodes (30 steps; episode 16628 of \cref{fig:s_roll_actl} is not counted). Under each screen: the goal, the last previous step instructions (\ldots: earlier ones), the annotator's instruction for this step (not shown to the models), the four drafts with their scores under $D$ (bold navy: kept; ties go to the higher mean token log-probability), the 8B alone and the gold action, with clicks on a 0--1000 scale; tick and cross marks: right and wrong. On the screens, rings mark the drafts' clicks, a peach $\times$ the 8B alone and a dark grey $+$ the gold click; in the middle column, drafts 1 and 4 click 9 units apart and the outer grey ring is draft 1. $\tau=0.5$.}
\label{fig:s_gallery_actl}
\end{figure}

\begin{figure}[!t]
\centering
\includegraphics[width=0.86\linewidth]{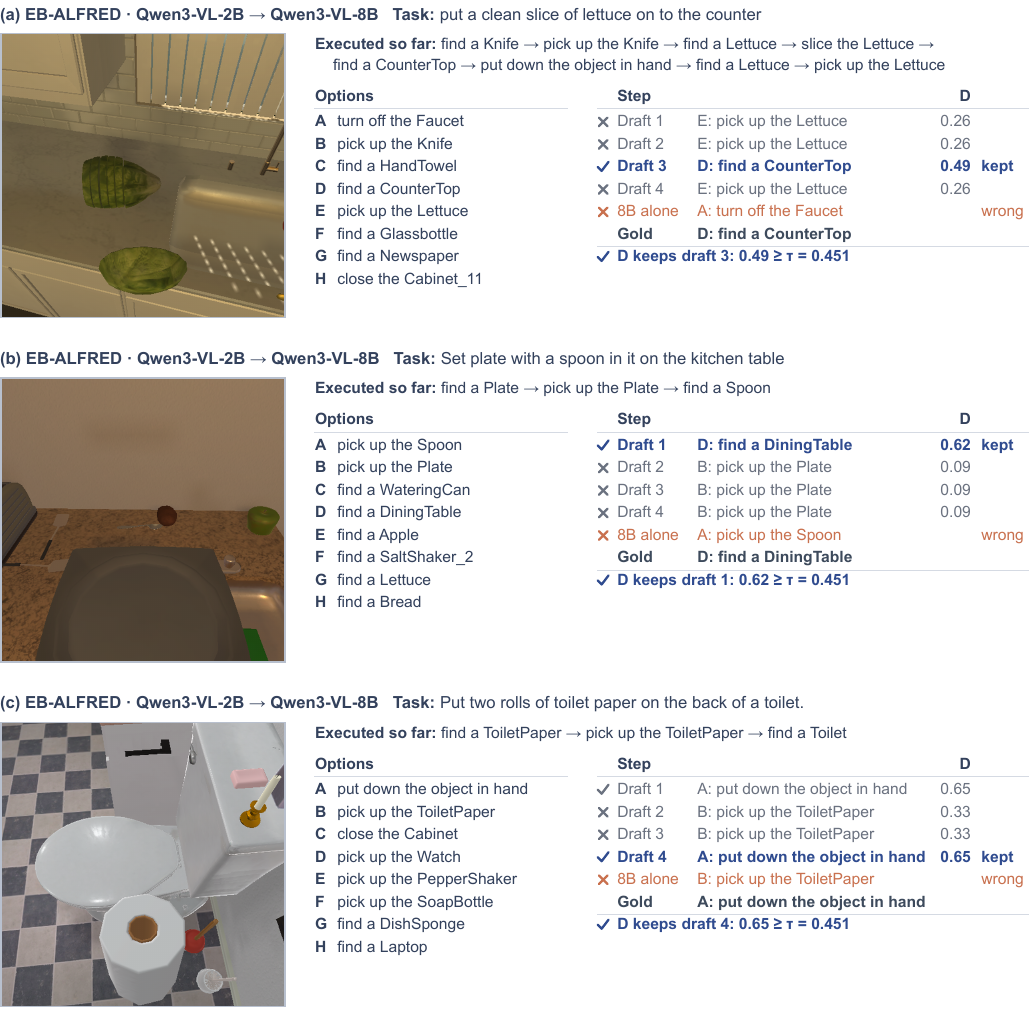}
\caption{\textbf{On held-out EB-ALFRED steps, $D$ finds the right action even when only one or two of the four 2B drafts propose it, so \method{} takes the right step where the 8B target alone does not.} A test step qualifies if the four drafts propose at least two different actions (drafts that cannot be parsed are not counted), \method{} keeps a right draft, and the target alone is wrong; we show the first such step of each of the first 3 qualifying episodes in test order, out of 48 qualifying episodes (67 steps; the two episodes shown in other figures are not counted). Left: the egocentric frame the models see. Right: the actions executed so far, the eight options, each draft's choice with its $D$ score, the 8B alone and the gold action; ticks and crosses mark right and wrong actions, and $\tau = 0.451$ is chosen on the dev split.}
\label{fig:s_gallery_eba}
\end{figure}

\begin{figure}[!t]
\centering
\includegraphics[width=0.86\linewidth]{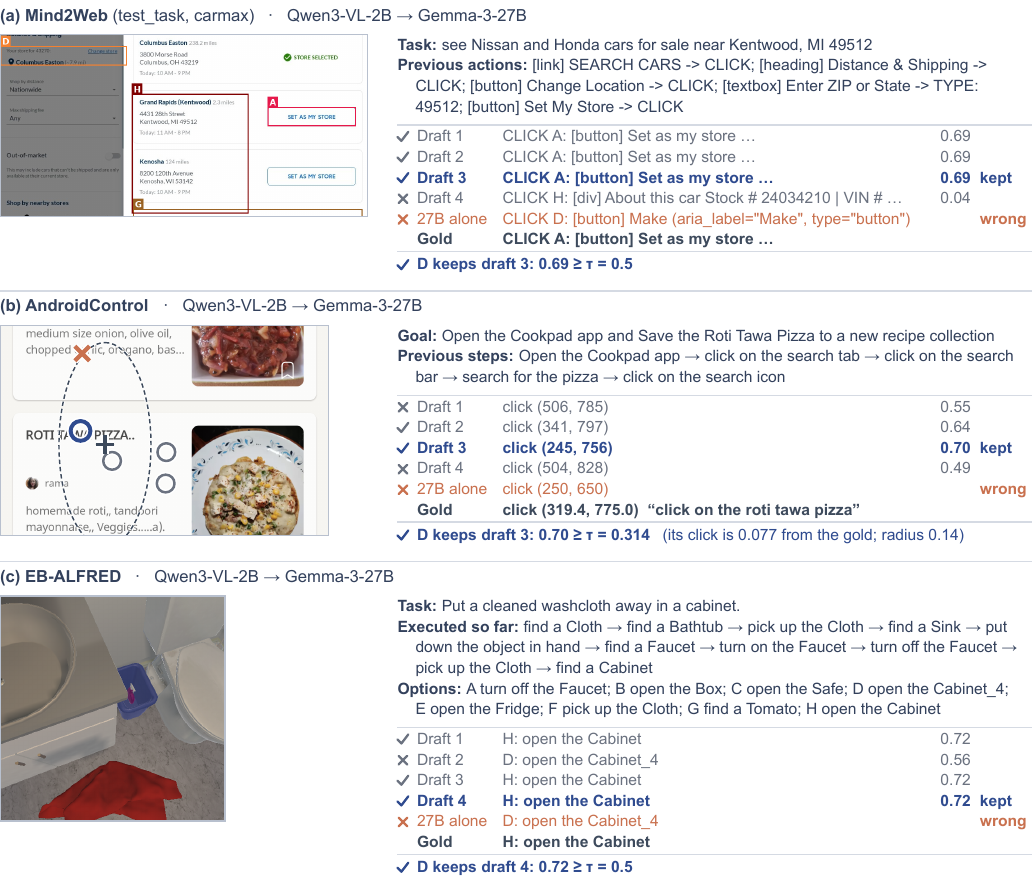}
\caption{\textbf{With a target from another family, $D$ still keeps a right Qwen3-VL-2B draft at steps where Gemma-3-27B alone is wrong.} Cases follow a fixed rule: a test step qualifies when the four drafts propose at least two different actions (clicks within the success radius count as one), \method{} keeps a right draft, and the target alone is wrong; in each cell we show the first such step of the first qualifying episode in test order that no other figure shows, out of 97 qualifying tasks (130 steps) on Mind2Web, 45 qualifying episodes (71 steps) on AndroidControl and 45 qualifying episodes (59 steps) on EB-ALFRED. Each row lists the four drafts with their $D$ scores (\checkmark{} right, $\times$ wrong; the kept draft in bold navy; ties in $D$ go to the draft with the higher mean log-probability), the action of Gemma-3-27B alone, the gold action, and $D$'s decision against the pair's dev-chosen threshold $\tau$. Mind2Web screenshots carry the Set-of-Mark boxes and letters shown to the models. On AndroidControl, rings mark the drafts' clicks, $\times$ the click of Gemma-3-27B alone, $+$ the gold click and the dashed outline the success radius (normalised distance 0.14); coordinates are on a 0--1000 scale, and the gold is printed with its annotated step instruction.}
\label{fig:s_gallery_g27}
\end{figure}

\begin{figure}[!t]
\centering
\includegraphics[width=0.86\linewidth]{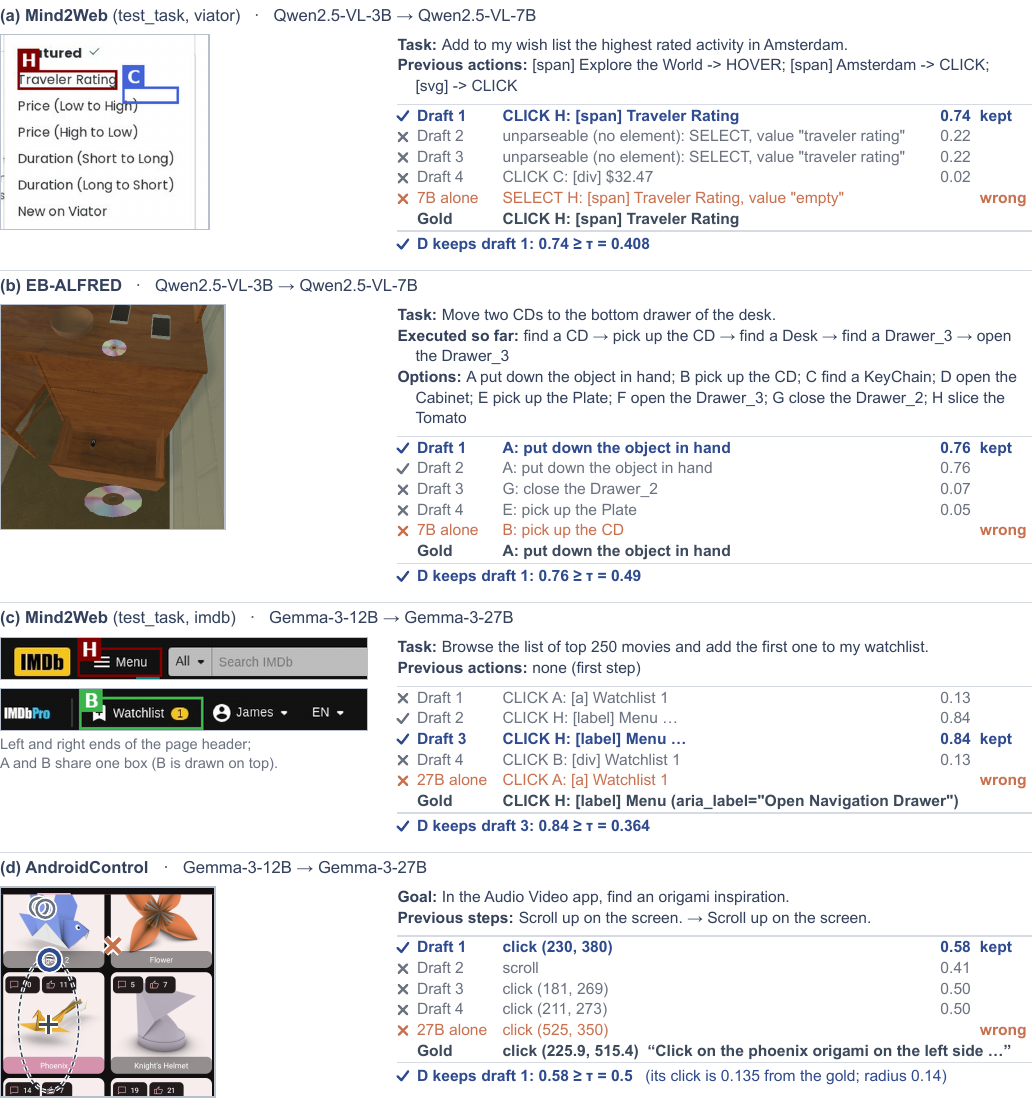}
\caption{\textbf{Within other model families too, $D$ keeps a right draft where the larger target alone is wrong, for Qwen2.5-VL-3B$\to$7B and Gemma-3-12B$\to$27B.} Cases follow the same fixed rule: a test step qualifies when the four drafts propose at least two different actions (clicks within the success radius count as one), \method{} keeps a right draft, and the target alone is wrong; in each cell we show the first such step of the first qualifying episode in test order that no other figure shows, out of (a) 127 qualifying tasks (215 steps), (b) 84 qualifying episodes (137 steps), (c) 94 qualifying tasks (126 steps) and (d) 28 qualifying episodes (36 steps). Rows as in \cref{fig:s_gallery_g27} (\checkmark{} right, $\times$ wrong; kept draft in bold navy; ties in $D$ go to the higher mean log-probability; $\tau$ is each pair's dev-chosen threshold). In (d), rings are the drafts' clicks, $\times$ the click of Gemma-3-27B alone, $+$ the gold click and the dashed outline the success radius (normalised distance 0.14).}
\label{fig:s_gallery_other}
\end{figure}

\Cref{fig:s_cases_win,fig:s_cases_fail,fig:s_cases_loop} show steps and episodes selected by the fixed rules stated after each of them.
The wins follow the pattern of \cref{sec:s_where}: the decision model scores the right element or the feasible action high and, where the drafts disagree, the alternatives low, and the target makes a different mistake.
\Cref{fig:s_gallery_m2w,fig:s_gallery_actl,fig:s_gallery_eba} add three test steps per agent benchmark, and \cref{fig:s_gallery_g27,fig:s_gallery_other} one step per cell for the targets and families of \cref{sec:s_families,sec:s_cross}, each chosen by the rule stated in its caption.

\FloatBarrier
The two steps of \cref{fig:s_cases_win} are selected by a fixed rule: (a) among the 116 Mind2Web test steps (83 tasks) on which the drafts propose at least two different actions, \method{} keeps a right draft and the 8B alone is wrong, the step on which $D$'s score of the kept draft exceeds its highest score of a wrong draft by the most; (b) among the 107 EB-ALFRED test steps on which \method{} keeps a right draft and the 8B alone takes an action that is impossible in the current state, the step with the highest $D$ score.

Without the screenshot, the decision model can no longer tell apart two fields that the text describes alike; with it, it keeps the draft that types the city into the departure field (\cref{fig:s_cases_fail}).
The step is selected by a fixed rule: among the 39 Mind2Web test steps (37 tasks) on which \method{} keeps a right draft with the screenshot and a wrong one without it ($\tau = 0.5$), the step on which, without the screenshot, $D$ scores the wrong draft highest relative to the right one.
\FloatBarrier

\FloatBarrier

In both episodes the 8B alone gets stuck and runs out of steps, whereas \method{} reaches the goal in 11 and 9 actions, handing a single decision to the 8B (\cref{fig:s_cases_loop}).
The episodes are selected by a fixed rule: on each task, among the test seeds 30--129 on which \method{} ($\tau = 0.3$) reaches the goal with at least one decision written by the 8B and the 8B alone does not (19 on GoToLocal, 26 on PickupLoc), ordered by \method{}'s number of actions (ties: lower seed), the seed at position $\lfloor N/3\rfloor + 1$ (7th and 9th).
\FloatBarrier

\FloatBarrier

In each of the three steps the drafts propose different elements, the decision model scores a right draft highest, and \method{} takes the right step where the 8B alone goes wrong (\cref{fig:s_gallery_m2w}).
\FloatBarrier

\FloatBarrier

In each of the three steps, the drafts propose different actions, the decision model singles out a right one, and \method{} keeps it where the 8B alone is wrong (\cref{fig:s_gallery_actl}).
\FloatBarrier

\FloatBarrier

Even when only one or two of the four drafts propose the right action, the decision model scores it highest, so \method{} takes the right step where the 8B alone does not (\cref{fig:s_gallery_eba}).
\FloatBarrier

\FloatBarrier

The pattern is the same with a target from another family: in all three cells, \method{} keeps a right Qwen3-VL-2B draft where Gemma-3-27B alone is wrong (\cref{fig:s_gallery_g27}).
\FloatBarrier

\FloatBarrier

Within the Qwen2.5-VL and Gemma-3 families too, the decision model keeps a right draft in each of the four cells where the larger target alone is wrong (\cref{fig:s_gallery_other}).
\FloatBarrier
\clearpage
\subsection{Rollouts}
\label{sec:s_rollouts}

\FloatBarrier

\begin{figure}[!htb]
\centering
\includegraphics[width=0.86\linewidth]{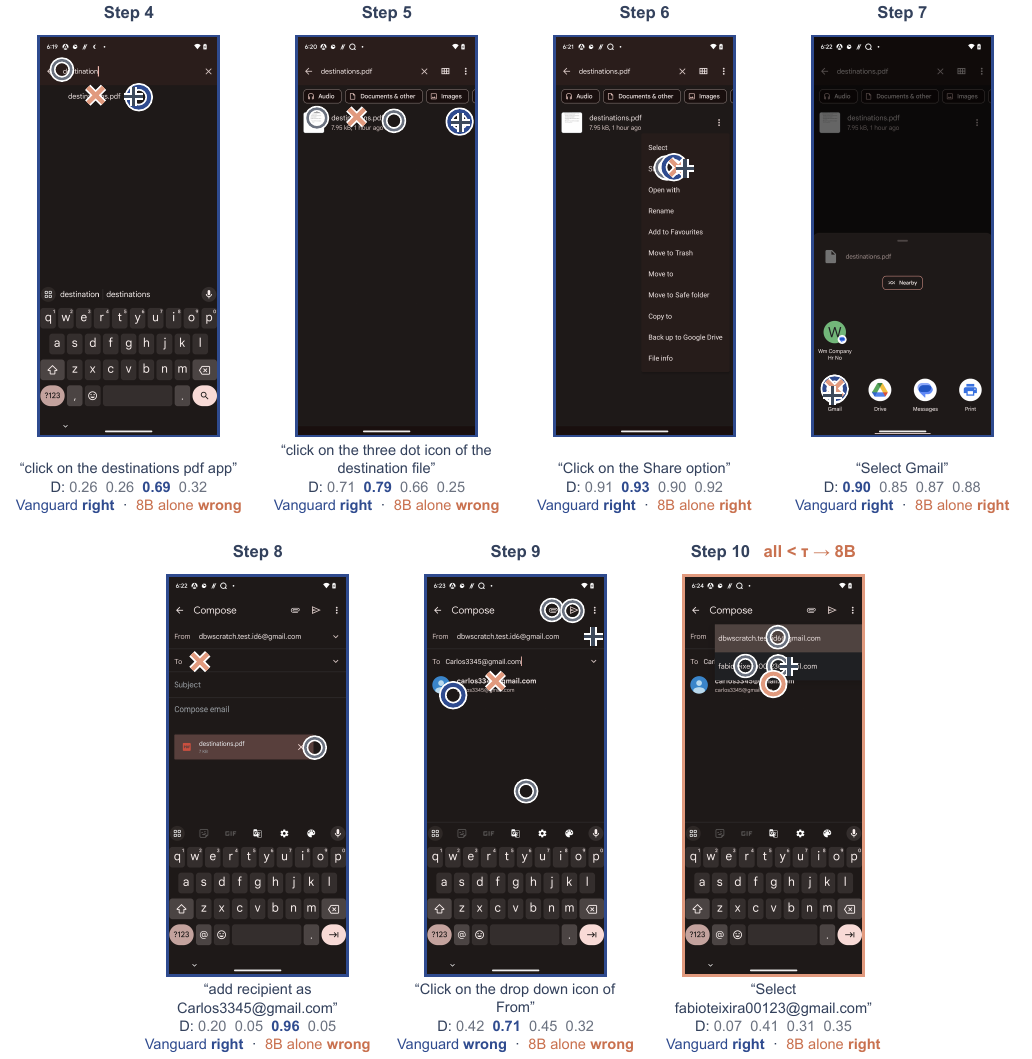}
\caption{\textbf{On a held-out AndroidControl episode, \method{} matches the annotated action on 6 of the 7 steps and the 8B alone on 3.} Steps 4 to 10 of test episode 16628 (goal: ``I want to send my friend a destination file via Gmail at Carlos3345@gmail.com in Files by Google App because he is in need of that file.''), each with the step's instruction and the scores of its four drafts under $D$ (bold navy: kept). On the screens, rings mark the drafts' clicks, a peach ring the click the 8B writes after every draft scored below $\tau=0.5$ (peach frame), a peach $\times$ the 8B alone and a dark grey $+$ the annotated click. On steps 4, 5 and 8 the kept draft is right and the 8B alone is wrong; on step 9 $D$ keeps a wrong draft (0.71) and the 8B alone is wrong as well; on step 10 every draft scores at most 0.41 and the 8B writes the step correctly. Selection: of the 98 held-out episodes, this one has the most steps on which \method{} is right and the 8B alone is wrong (4, at steps 4, 5, 8 and 11; no other episode has more than 3), and we show the seven steps from the first of them on.}
\label{fig:s_roll_actl}
\end{figure}

\Cref{fig:s_roll_actl,fig:s_roll_eba} follow one held-out episode each, on AndroidControl and on EB-ALFRED, step by step, with the screen or egocentric frame the agent saw, the four drafts and their scores under $D$, the decision and the target model alone. Both episodes are chosen by the same rule, stated in the captions. As in the main evaluation, each step is judged on the recorded screen or frame with the reference history.

Over the seven steps, \method{} matches the annotated action on six, against three for the 8B alone; on the one step where every draft scores below the threshold, the 8B writes the step and gets it right (\cref{fig:s_roll_actl}).

\FloatBarrier

\begin{figure}[!t]
\centering
\includegraphics[width=0.86\linewidth]{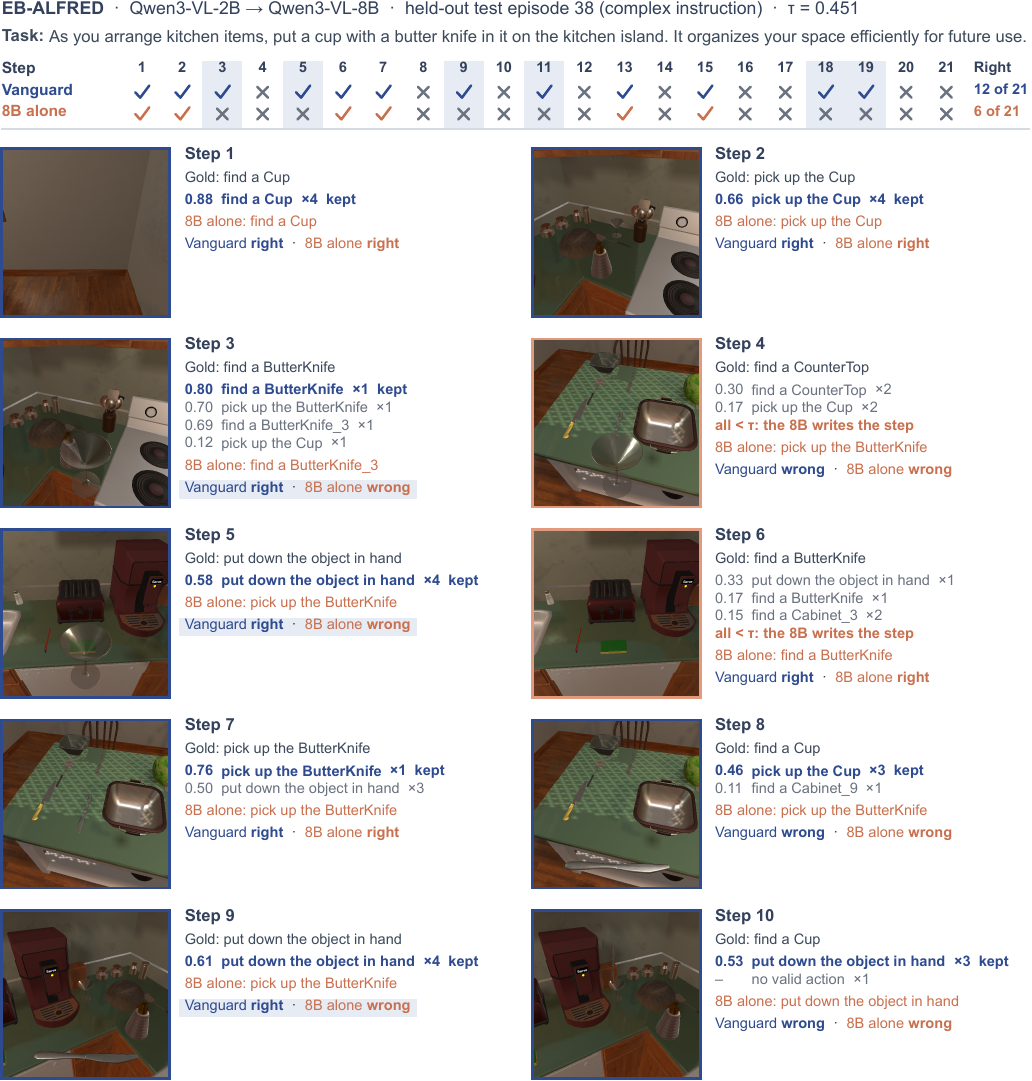}
\caption{\textbf{On a held-out EB-ALFRED episode, \method{} matches the gold action on 12 of the 21 steps and the 8B alone on 6.} All 21 steps of test episode 38 of the complex-instruction split (Qwen3-VL-2B$\to$Qwen3-VL-8B, $\tau=0.451$), steps 1 to 10 on this page and 11 to 21 on the next. Top: the task and, for every step, whether \method{} and the 8B alone are right (\checkmark) or wrong ($\times$); shaded: \method{} right and the 8B alone wrong. Each step shows the egocentric frame the agent saw, the gold action (from the dataset's shortest successful trajectory, which our evaluation follows), the distinct actions of the four drafts with their scores under $D$ and how many drafts propose each (bold navy: kept; --: a draft that cannot be parsed; ties in $D$ go to the higher mean token log-probability), and the action of the 8B alone; the frame is navy when \method{} keeps a draft and peach when every draft scores below $\tau$ and the 8B writes the step. Selection: of the 182 held-out episodes, 82 have at least one step on which \method{} is right and the 8B alone is wrong (132 such steps in all); this one has the most (6, at steps 3, 5, 9, 11, 18 and 19; no other episode has more than 5).}
\label{fig:s_roll_eba}
\end{figure}
\FloatBarrier

Over the 21 steps, \method{} matches the gold action twice as often as the 8B alone (12 against 6), and on no step is the 8B alone right where \method{} is wrong (\cref{fig:s_roll_eba}).
\FloatBarrier

\begin{figure}[!t]
\ContinuedFloat
\centering
\includegraphics[width=0.86\linewidth]{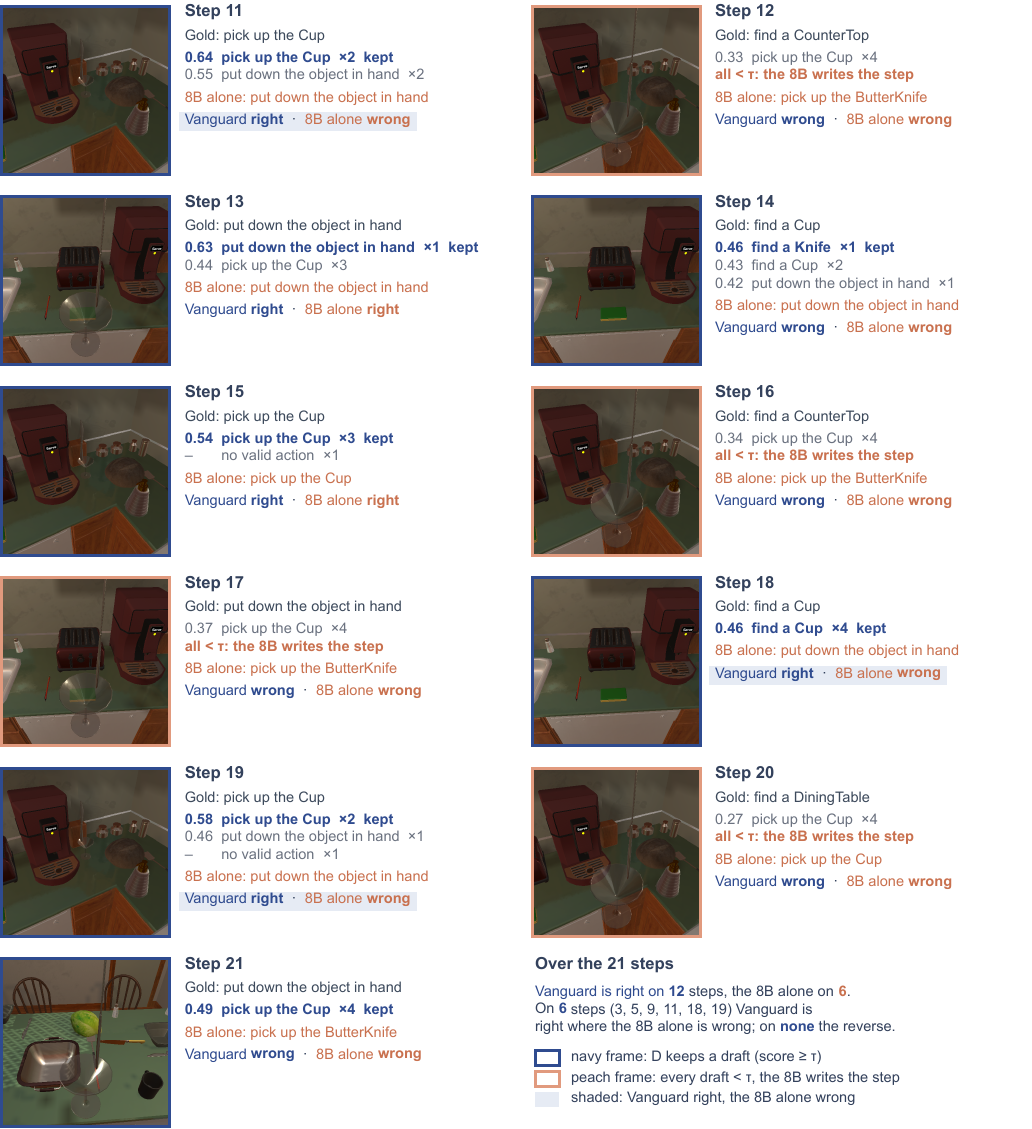}
\caption{(continued) Steps 11 to 21 of the EB-ALFRED episode of the previous page, shown in the same way.}
\end{figure}
\FloatBarrier

In the second half of the episode, \method{} is right and the 8B alone wrong at steps 11, 18 and 19.
\FloatBarrier

\FloatBarrier

\end{document}